\documentclass{article}

\usepackage[main, final]{neurips_2026}

\usepackage[utf8]{inputenc} 
\usepackage[T1]{fontenc}    
\usepackage{hyperref}       
\usepackage{url}            
\usepackage{booktabs}       
\usepackage{amsfonts}       
\usepackage{nicefrac}       
\usepackage{microtype}      

\usepackage{amssymb}
\usepackage{amsmath}
\usepackage{wrapfig}

\def\eg{\emph{e.g.}}

\def\etc{\emph{etc.}}

\usepackage{graphicx}

\usepackage{multirow}
\usepackage[table]{xcolor}  

\newcommand{\categorycell}[1]{\multicolumn{2}{l}{\cellcolor{gray!10} #1}}
\usepackage{tcolorbox}                                       
\usepackage{enumitem}

\definecolor{correct}{RGB}{34,139,34}
\definecolor{wrong}{RGB}{200,30,30}

\newcommand{\errorcase}[4]{%
  \begin{minipage}[t]{0.235\linewidth}
    \centering
    \includegraphics[width=\linewidth,height=2.4cm,keepaspectratio]{#1}\\[1pt]
    {\tiny \textbf{Q:} #2}\\[1pt]
    {\tiny \textcolor{correct}{\textbf{GT:} #3} \; \textcolor{wrong}{\textbf{Pred:} #4}}
  \end{minipage}%
}

\title{HoloCount: A Holistic Visual Counting Benchmark for MLLMs}

\author{%
  Jinhong Deng
    \\
  Meituan\\
  \texttt{dengjinhong@meituan.com} \\
  \And
  Limeng Qiao \\
  Meituan \\
  \texttt{qiaolm@pku.edu.cn} \\
  \And
  Guanglu Wan \\
  Meituan \\
  \texttt{wanguanglu@meituan.com}
}

\begin{document}

\maketitle

\begin{abstract}
Visual counting is a fundamental pillar of multimodal intelligence, requiring a seamless integration of fine-grained grounding and spatial reasoning. While Multimodal Large Language Models (MLLMs) have achieved remarkable success in qualitative scene understanding, their quantitative precision remains a significant bottleneck, often characterized by persistent numerical hallucinations. Existing counting benchmarks primarily focus on basic perception in simplified contexts, failing to capture the complex failure modes that emerge under logical constraints or adversarial conditions. To address these limitations, we introduce HoloCount, a holistic and diagnostically rich benchmark structured around a three-level hierarchical taxonomy. HoloCount evaluates MLLMs across: (1) Semantic Counting, focusing on atomic and property-based enumeration; (2) Analytical Counting, assessing logical composition through spatial and set-based reasoning; and (3) Robustness Testing, probing model integrity against adverse scenarios and grounded counter-priors, such as high-density scenes and linguistic biases. Through an exhaustive evaluation of over 20 state-of-the-art MLLMs, we reveal a critical performance gap: even top-tier models degrade significantly as tasks transition from perception to complex analytical reasoning and adverse scenarios. Our findings provide a systematic landscape of current MLLM counting capabilities and offer a roadmap for developing more grounded and reliable multimodal systems. The dataset is available at \url{https://mm-mvr.github.io/HoloCount/}.
\end{abstract}

\section{Introduction}
\label{sec:intro}

Visual counting~\cite{tamarapalli2025countqa,countbench,pixmo}, the ability to quantify instances of specific categories in an image, is an important skill that reflects a key aspect of visual intelligence. For modern Multimodal Large Language Models (MLLMs)~\cite{LLaVA,bai2025qwen3-vl,wang2024qwen2-vl,team2025kimi-vl,keye-vl,huang2026step3}, counting is not merely a numerical task but a rigorous test of fine-grained grounding~\cite{gdino,visual_grounding} and spatial reasoning~\cite{cheng2024spatialrgpt}. While MLLMs excel at qualitative scene understanding such as Visual Question Answering (VQA)~\cite{zhang2024mme}, their limited quantitative precision remains a critical barrier for applications such as logistics~\cite{woschank2020review}, safety monitoring~\cite{nath2020deep}, and retail analytics~\cite{abed2024convolutional}, where numerical errors carry high costs.

Despite the rapid evolution of MLLMs~\cite{LLaVA,bai2025qwen3-vl,wang2024qwen2-vl,team2025kimi-vl} such as the GPT series~\cite{achiam2023gpt} and the Qwen-VL series~\cite{bai2025qwen3-vl,wang2024qwen2-vl}, their ability to perform reliable counting remains surprisingly fragile~\cite{countbench,tamarapalli2025countqa}. Although these models exhibit strong zero-shot performance in general scene description, they often struggle with precision when tasked with enumerating objects. This counting limitation frequently manifests as numerical hallucinations, where the model confidently reports a count that bears little resemblance to the true quantity, undermining reliability in count-sensitive environments. Recent benchmarking efforts~\cite{countbench,pixmo,tamarapalli2025countqa}, including CountBench~\cite{countbench} and CountQA~\cite{tamarapalli2025countqa}, have begun to address this challenge but remain constrained by limited object diversity and a focus on simple perceptual tasks. They often fail to reveal complex failure modes that arise when models face property constraints, logical constraints, or adversarial visual conditions.

\begin{figure}
    \centering
    \includegraphics[width=\linewidth]{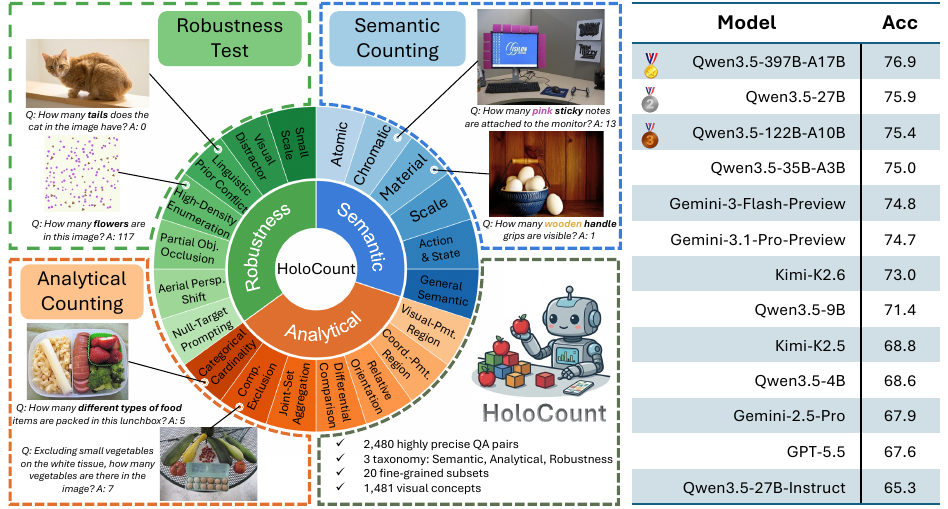}
    \caption{Taxonomy overview of the HoloCount benchmark. The dataset features three taxonomy containing 2,480 QA pairs across 20 fine-grained counting tasks.}
    \label{fig:teaser}
\end{figure}

To address these gaps, we introduce HoloCount, a comprehensive benchmark designed to rigorously evaluate the counting competence of MLLMs. Unlike prior benchmarks~\cite{zhang2016single,pixmo,countbench} (see detailed comparison in Table~\ref{tab:benchmark_comparison}), HoloCount is organized around a three-level hierarchical taxonomy that shifts the focus from mere superficial perception in idealized settings to the advanced cognitive processing and environmental resilience required for true numerical grounding. These three distinct pillars evaluate equally vital facets of counting competence: (1) \textbf{Semantic Counting} assesses foundational skills through atomic counting (basic enumeration) and property-based filtering, testing the model's ability to isolate target objects by fine-grained attributes such as object chromatic and material properties. (2) \textbf{Analytical Counting} examines the interplay of vision and logic through logical composition, requiring models to execute spatial-based reasoning, set-based operations, and conceptual calculus, such as multi-set aggregation, exclusions, and relational comparisons. (3) \textbf{Robustness Testing} probes the systemic limits of model reliability under adverse perceptual domains (\textit{e.g.}, heavy occlusion, high object density, and aerial perspectives) while explicitly evaluating the model's resistance to deceptive counter-priors, including linguistic biases, null targets, and visual illusions. To realize this multi-dimensional diagnostic pipeline, HoloCount comprises a meticulously curated dataset of 2,480 highly precise QA pairs partitioned across 20 fine-grained subsets, as shown in Fig.~\ref{fig:teaser}. By structuring this extensive scale around distinct cognitive axes rather than a flat, unstructured mixture of queries, our benchmark provides a decoupled evaluation framework. Consequently, HoloCount can comprehensively review the deep-seated counting vulnerabilities of MLLMs, systematically isolating whether a model's failure stems from a visual alignment oversight, a logical reasoning flaw, or a fragile reliance on superficial patterns.

\begin{table*}[t]
\centering
\caption{Comprehensive comparison between HoloCount and existing visual counting benchmarks.}
\label{tab:benchmark_comparison}
\resizebox{1.0\linewidth}{!}{
    \begin{tabular}{lcccccc}
    \toprule
    \textbf{Benchmark} & \textbf{\# Samples} & \textbf{\# Visual Concepts} & \textbf{Taxonomy} & \textbf{Tasks} & \textbf{Analytical} & \textbf{Robustness} \\ \midrule
    ShanghaiTech~\cite{zhang2016single} & 1,198  & 1  & No  & 1 & $\times$  & $\times$ \\
    NWPU~\cite{wang2020nwpu} &  5,109  & 1  & No  & 1 & $\times$  & $\times$ \\
    JHU-CROWD++~\cite{jhu_crowd} & 4,372  & 1  & No  & 1 & $\times$  & $\times$ \\
    CARPK~\cite{hsieh2017drone} & 1,148  & 1  & No  & 1 & $\times$  & $\times$ \\
    UCF QNRF~\cite{idrees2018composition} & 1,535  & 1  & No  & 1 & $\times$  & $\times$ \\
    FSC-147~\cite{fsc147}  &  6,135  & 147  & No  & 1 & $\times$  & $\times$ \\ \hline
    CountBench~\cite{countbench}        & 488  & 231  & No & 1 & $\times$  & $\times$ \\
    PixMo-Count~\cite{pixmo}        & 540 & 60 & No & 1 & $\times$  & $\times$ \\
    CountQA~\cite{tamarapalli2025countqa}& 1,528  & 534  & No & 1 & $\times$  & $\times$ \\ \hline
    \textbf{HoloCount (Ours)} & \textbf{2,480} & \textbf{1,481} & \textbf{Yes} & \textbf{20} & \checkmark & \checkmark \\ \bottomrule
    \end{tabular}
}
\end{table*}

Through an exhaustive evaluation of over 20 leading MLLMs~\cite{wang2024qwen2-vl,bai2025qwen3-vl}, we uncover a sobering reality: model performance deteriorates markedly as tasks progress from basic perception to analytical reasoning and robustness challenges. The experimental results also reveal valuable insights and provide a clear roadmap for enhancing the visual counting capabilities of MLLMs.

Our contributions are summarized as follows:
\begin{itemize}
\item \textbf{The HoloCount Benchmark:} We present HoloCount, a novel benchmark designed to rigorously evaluate the counting competence of MLLMs. Unlike prior flat benchmarks, HoloCount is organized around a hierarchical three-level taxonomy—Semantic Counting, Analytical Counting, and Robustness Testing—that shifts focus from superficial perception to advanced cognitive processing and environmental resilience.
\item \textbf{Fine-grained Tasks:} We construct a meticulously curated dataset of 2,480 high-precision QA pairs spanning 20 fine-grained counting tasks and 1,481 visual concepts, making it the most diagnostically rich counting benchmark to date in terms of task diversity, visual concepts, and structured evaluation axes.
\item \textbf{Extensive Evaluation \& Actionable Insights:} We conduct an exhaustive empirical study of state-of-the-art models (including the Qwen-VL series, the GPT series, \etc), revealing that performance deteriorates markedly from basic perception to analytical reasoning and robustness challenges, thereby providing a sobering reality check on current MLLM counting capabilities. The benchmark and evaluation framework identify root causes of model failures, offering a clear roadmap for developing truly grounded, numerically reliable systems.

\end{itemize}

\section{Related Work}
\label{sec:related_work}

\noindent \textbf{Visual Counting in VLMs.} Traditional visual counting research~\cite{zhang2016single,count_survey2,jhu_crowd} was largely confined to specialized domains with fixed, narrow categories, such as crowd counting~\cite{zhang2016single} or cell enumeration~\cite{bcdata,vgg_cell_counting}. Early methodologies often relied on task-specific architectures, utilizing normalized attention weights~\cite{zhang2018learning} or relation networks~\cite{acharya2019tallyqa} to infer object quantities. With the emergence of Large Vision-Language Models (VLMs)~\cite{siglip}, research shifted toward leveraging the zero-shot capabilities of foundational models like CLIP~\cite{clip} to count objects in an open-vocabulary manner~\cite{countbench,tamarapalli2025countqa}. However, despite their strong contrastive alignment, these dual-encoder models often struggle to parse complex semantics or reason through free-form counting queries. The transition to Multimodal Large Language Models (MLLMs) has significantly expanded the scope of counting, allowing for more conversational and multi-step interactions. Recent works~\cite{pixmo} have attempted to improve these capabilities through targeted data and architectural innovations. For instance, PixMo (Pixels for Molmo)~\cite{pixmo} introduced a novel 2D pointing dataset that allows models to "count by pointing," significantly enhancing grounding and numerical accuracy through point-based supervision.

\noindent \textbf{Counting Benchmarks.} Early visual counting benchmarks were predominantly domain-specific, focusing on high-density scenarios such as crowd analysis (\textit{e.g.}, ShanghaiTech~\cite{zhang2016single}, NWPU-Crowd~\cite{wang2020nwpu}) or vehicle monitoring (\textit{e.g.}, CARPK~\cite{hsieh2017drone}). These datasets typically formulate counting as a density map regression task restricted to a closed set of categories. To achieve category-agnostic counting, few-shot benchmarks such as FSC-147~\cite{fsc147} and FSCD-LVIS~\cite{nguyen2022few} were introduced, requiring models to count arbitrary objects given a few visual exemplars. Some works~\cite{rong2026unicbenchunifiedcountingbenchmark,lu2025avreasoner,tsuchiya2026ecbench,liu2026momentvideodiagnosingtemporalfidelity} focus on counting tasks in more modalities such as audio and video. With the emergence of MLLMs, the paradigm has shifted toward open-vocabulary counting guided by natural language instructions~\cite{liu2026countgranularity,countbench,tamarapalli2025countqa,pixmo,nguyen2025currentaimodelscount}. CountBench~\cite{countbench} and CountQA~\cite{tamarapalli2025countqa} pioneered this space by evaluating zero-shot counting via VQA formats, while PixMo-Count~\cite{pixmo} utilized point-based supervision to improve numerical grounding. However, these benchmarks largely focus on basic perception and lack the depth to diagnose failures in complex reasoning or robustness under deceptive conditions. In this work, HoloCount bridges this gap by introducing a three-level hierarchical evaluation that systematically decouples semantic, analytical, and adverse robustness.

\section{The HoloCount Benchmark}
\label{sec:HoloCount}

\subsection{Task Formulation}

The objective of visual counting within the framework of Multimodal Large Language Models (MLLMs) is to quantify a specific subset of objects in an image $\mathcal{I}$ based on a natural language instruction $\mathcal{Q}$, \eg, ``\textit{How many cats are there in the image}''. Formally, given the input pair $(\mathcal{I}, \mathcal{Q})$, the model $\mathcal{M}$ aims to generate a textual response $\mathcal{A}$ that contains the predicted count $\hat{N}$. This process is defined as:
\begin{equation}
    \mathcal{A} = \mathcal{M}(\mathcal{I}, \mathcal{Q}; \theta)
\end{equation}
where $\theta$ represents the model parameters. The success of the task is measured by the alignment between the extracted numerical value $\hat{N} \in \mathcal{A}$ and the ground-truth cardinality $N$.

\begin{figure}
    \centering
    \includegraphics[width=1.0\linewidth]{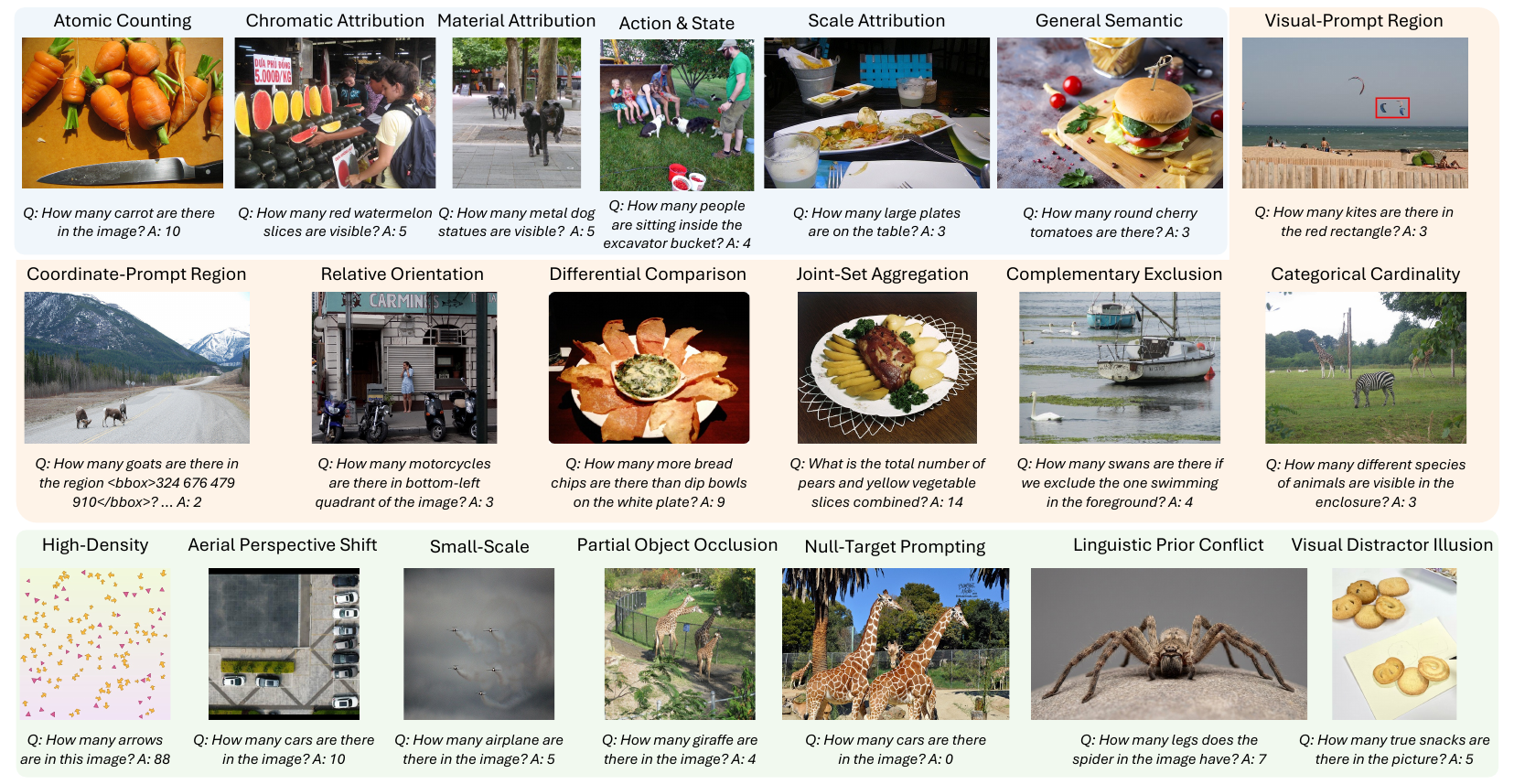}
    \caption{Example tasks from the HoloCount benchmark, which comprises 20 fine-grained subtasks.}
    \label{fig:example}
\end{figure}

\subsection{Overview}

Traditional counting benchmarks~\cite{zhang2016single,count_survey2,jhu_crowd} typically evaluate counting models on a limited set of fixed visual concepts, such as vehicle monitoring in CARPK~\cite{hsieh2017drone}, where models count cars in parking lots. With the emergence of MLLMs, many VQA benchmarks~\cite{zhang2024mme} have been proposed to evaluate the visual understanding capability of these models, including a few examples of counting questions. Recently, CountBench~\cite{countbench} and CountQA~\cite{tamarapalli2025countqa} have been specifically proposed to focus on the counting problem. However, these benchmarks remain limited by narrow object diversity (see comparison of visual concepts in Table~\ref{tab:benchmark_comparison}) and a focus on simple perception. They often fail to capture the complex failure modes that occur when models are subjected to property constraints, logical constraints, or adversarial visual conditions. Most importantly, these benchmarks adopt a flat structure; the evaluation only yields a single accuracy score, providing no insight into where or why the model fails. Consequently, they cannot offer effective guidance for improving the counting capability of the model.

\begin{table}[tbp]
\centering
\caption{Statistics of the HoloCount benchmark: number of QA pairs per subset and total.}
\label{tab:dataset_stats}
\resizebox{0.9\linewidth}{!}{
\begin{tabular}{lrlr}
\toprule
\textbf{Subsets} & \textbf{Number} & \textbf{Subsets} & \textbf{Number} \\
\midrule
\categorycell{Semantic Counting} &&  \\
~~~~$\cdot$ Atomic Counting (Atom.)               & 182 & ~~~~$\cdot$ Joint-Set Aggregation (Aggr.)       & 100 \\
~~~~$\cdot$ Chromatic Attribution (Chrom.)        & 202 &  ~~~~$\cdot$ Complementary Exclusion (Excl.)      & 101 \\
~~~~$\cdot$ Material Attribution (Mater.)         & 101 & ~~~~$\cdot$ Categorical Cardinality (Categ.)     & 101 \\
~~~~$\cdot$ Scale Attribution (Scale)            & 100 & \categorycell{Robustness Testing} \\
~~~~$\cdot$ Action \& State Attribution (Act.\&St.)  & 102 & ~~~~$\cdot$ High-Density Enumeration (Dense)     & 100 \\
~~~~$\cdot$ General Semantic Attribution (Gen.)  & 101 & ~~~~$\cdot$ Aerial Perspective Shift (Aerial)     & 109 \\
\categorycell{Analytical Counting} &  ~~~~$\cdot$ Small-Scale Enumeration (small)      & 101 \\
~~~~$\cdot$ Visual-Prompt Region Grounding (Vis.) & 102 & ~~~~$\cdot$ Partial Object Occlusion (Occl.)     & 103 \\
~~~~$\cdot$ Coordinate-Prompt Region Grounding (Coord.) & 78  &~~~~$\cdot$ Null-Target Prompting (Null.T.)        & 250 \\
~~~~$\cdot$ Relative Canonical Orientation (Orient.) & 84& ~~~~$\cdot$ Linguistic Prior Conflict (Ling.P.)    & 163 \\
~~~~$\cdot$ Differential Comparison (Diff.) & 100& ~~~~$\cdot$ Visual Distractor Illusion (Vis.D.)   & 200 \\

\midrule
\multicolumn{2}{l}{\textbf{Total}} & \multicolumn{2}{l}{\textbf{2480}} \\
\bottomrule
\end{tabular}
}
\end{table}

To bridge this gap, we propose HoloCount, a comprehensive and rigorous benchmark designed to holistically assess the counting ability of MLLMs. HoloCount is structured around a three-level hierarchical taxonomy that shifts the focus from perception to cognitive counting ability and adverse scenarios, including: (1) \textbf{Semantic Counting} assesses foundational skills through atomic counting (basic enumeration) and property-based filtering, testing the model's ability to isolate target objects by fine-grained attributes such as object chromatic and material properties. (2) \textbf{Analytical Counting} examines the interplay of vision and logic through logical composition, requiring models to execute spatial-based reasoning, set-based operations, and conceptual calculus, such as multi-set aggregation, exclusions, and relational comparisons. (3) \textbf{Robustness Testing} probes the systemic limits of model reliability under adverse perceptual domains (\textit{e.g.}, heavy occlusion, high object density, and aerial perspectives) while explicitly evaluating the model's resistance to deceptive counter-priors, including linguistic biases, null targets, and visual illusions. In particular, HoloCount contains 20 subsets and 2,480 QA samples. As shown in Table~\ref{tab:dataset_stats}, Semantic Counting includes 6 subsets covering atomic counting, chromatic, material, scale, action and state attributions. Analytical Counting evaluates models from the perspectives of spatial-based reasoning (Visual-Prompt Region Grounding, Coordinate-Prompt Region Grounding, Relative Canonical Orientation) and set-based logic reasoning (Differential Comparison, Joint-Set Aggregation, Complementary Exclusion, Categorical Cardinality). Robustness Testing focuses on adverse conditions (High-Density Enumeration, Aerial Perspective Shift, Small-Scale Enumeration, Partial Object Occlusion) and hallucination (Null-Target Prompting, Linguistic Prior Conflict, Visual Distractor Illusion). We also present example tasks in Fig.~\ref{fig:example} for better understanding of the HoloCount benchmark. HoloCount not only provides a more diverse set of evaluations but also serves as an effective tool for analyzing the failure modes of counting models.

\subsection{Data Curation Process}

\begin{figure}
    \centering
    \includegraphics[width=1.0\linewidth]{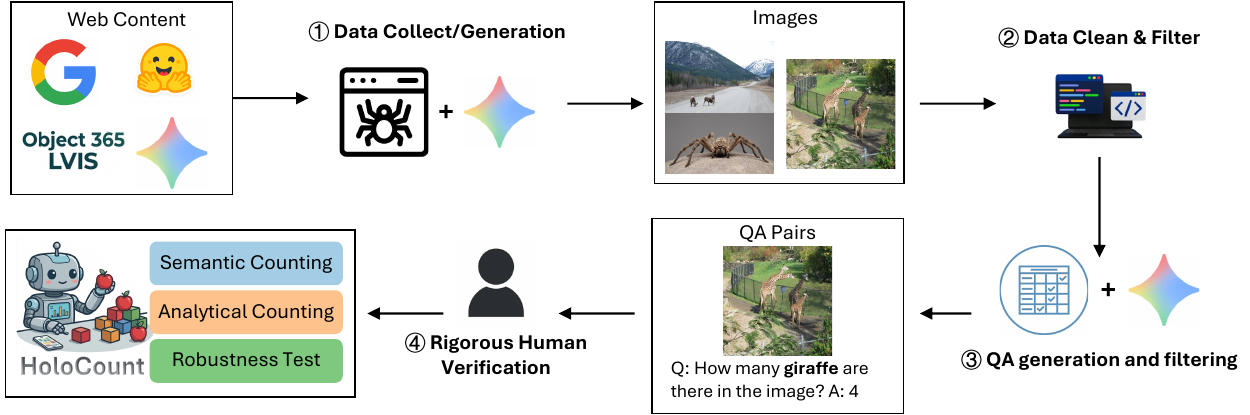}
    \caption{The general data curation pipeline of the HoloCount benchmark.} 
    \label{fig:data_pipeline}
\end{figure}

To construct a high-quality and reliable benchmark, we adhere to two core principles during the development of HoloCount: (1) \textbf{Ambiguity Elimination}: Visual counting often suffers from semantic or boundary ambiguity (\eg, whether to count a highly occluded object). We prune samples with vague instructions or ambiguous visual evidence, ensuring that each query has a singular, well-defined numerical answer.
(2) \textbf{Category and Scenario Diversity}: We aim for maximum coverage across diverse object categories and environmental contexts to test the true generalization of MLLMs.

\textbf{General Data Pipeline.} To ensure the diversity, difficulty, and diagnostic depth of HoloCount, we implement a multi-stage data curation pipeline, transitioning from large-scale collection to rigorous human-in-the-loop verification. The general data pipeline is shown in Fig.~\ref{fig:data_pipeline} and contains four core steps:
(1) \textbf{Data Collection/Generation}. To meet different evaluation purposes and ease the data construction process, we adaptively collect images from the internet, detection datasets, specialized benchmarks, and manual synthesis for specific purposes. (2) \textbf{Data Cleaning and Filtering}. We exclude images with extreme sizes or aspect ratios. Moreover, we also exclude images where target objects occupy extreme scales (too small for clear identification or too large to be reliably perceived). (3) \textbf{QA Generation and Filtering}. We use advanced MLLMs to generate QA pairs and employ another model to check the correctness of the generated QA pairs. (4) \textbf{Rigorous Human Verification}. The final and most critical stage involves exhaustive human audit. Each data sample undergoes manual review by expert annotators. The annotators are tasked with correcting any numerical inaccuracies and ensuring that the queries are linguistically natural and unambiguous.

Note that the data pipeline described above represents the general strategy for HoloCount benchmarks. Some steps may be deliberately broken for specific subsets to serve particular evaluation purposes. For example, we intentionally retain small visual objects in the subset of Small-Scale Enumeration dedicated to small object counting. We will provide detailed introductions for these subset-specific designs later.

\subsubsection{Semantic Counting}
Semantic Counting aims at evaluating whether the model has the ability to bind the semantics of objects in an image (see Fig.~\ref{fig:example} for example). Existing counting benchmarks belong to this taxonomy.

\noindent \textbf{Atomic Counting}. This is the most common counting problem, involving basic concepts such as ``how many apples are there in the image?'' This subset is very close to CountBench~\cite{countbench}, PixMo Count~\cite{pixmo}, and CountQA~\cite{tamarapalli2025countqa}. To construct this subset, we first sample images from the validation set of LVIS~\cite{lvis}. We constrain the maximum count number to 20 and filter out samples where the objects occupy an area of the image less than $0.1\%$ or more than $85.0\%$. Since detection datasets often contain miss-annotated objects that significantly influence the counting training signal, we utilize MLLMs to first check the correctness of the annotations. Then, human experts double-check the final annotations to ensure the quality of the dataset.

\noindent \textbf{Attribution: Chromatic, Material, Scale, Action \& State, and General Semantic}. Attributes represent core semantic dimensions for counting. In this work, we consider chromatic, material, size, action and state, and other general semantic attributes as key factors for visual object counting. To construct attribute-based counting QA pairs, we prompt a strong MLLM, Gemini-3-Flash-Preview, to generate candidate questions and answers conditioned on each attribute type. We then employ another model as a verifier to check the correctness of the generated QA pairs, filtering out inconsistent or invalid ones. Finally, we conduct manual verification by human annotators to ensure high quality and reliability of the final QA dataset.

\subsubsection{Analytical Counting}

Existing counting benchmarks~\cite{countbench,pixmo,tamarapalli2025countqa} primarily assess perceptual counting, i.e., directly enumerating objects of a given category. However, real-world visual reasoning often requires more than direct perception: it demands analytical counting, where the model must integrate visual information with logical operations, spatial constraints, or set-based reasoning. For example, answering ``How many more red blocks than blue blocks?'' involves not only counting each color but also performing a relational comparison. Conventional benchmarks fail to evaluate such cognitive counting abilities because they assume a flat, direct mapping from image to count. They neither introduce compositional logic nor require relational thinking. As a result, a model that performs well on atomic counting may still fail catastrophically when faced with analytical counting tasks---a failure mode that flat accuracy scores cannot reveal. HoloCount explicitly targets this gap by introducing \textbf{Analytical Counting}, which comprises two broad categories: (i) spatial-based reasoning and (ii) set-based logical reasoning.

\noindent\textbf{Spatial-based Analytical Counting.} Spatial-based analytical counting requires the model to count objects within a region of interest (ROI) that is defined not by object categories but by spatial reference. This tests the model's ability to ground counting instructions to specific visual locations. We design three subsets: (1) Visual-Prompt Region Grounding: The ROI is marked by a visual prompt (\textit{e.g.}, a bounding box). The model must count only the objects falling inside that region, ignoring visually similar distractors outside. This evaluates visuospatial alignment without textual coordinates. (2) Coordinate-Prompt Region Grounding: The ROI is specified via numerical coordinates (\textit{e.g.}, ``in the region <bbox>x\_min y\_min x\_max y\_max</bbox>''). The model must translate symbolic spatial descriptions into pixel-level counting. This tests the integration of language-based geometry with visual enumeration. (3) Relative Canonical Orientation: The ROI is defined relative to object orientations (\textit{e.g.}, ``in the bottom-left quadrant of the image''). This requires the model to recognize canonical directions and apply them as a spatial filter. These tasks are absent from prior counting benchmarks, which either assume global counting or rely on trivial region definitions. By contrast, HoloCount forces models to jointly perform spatial reasoning and enumeration, revealing weaknesses in grounding and attention. Obviously, the construction of visual grounded counting requires the help of bounding boxes of objects. We therefore start from LVIS~\cite{lvis} and utilize an MLLM to check the annotations. Then, we build the three subsets according to the bounding boxes. After that, human annotators further check the generated QA pairs.

\noindent\textbf{Set-based Analytical Counting.} Set-based analytical counting involves operations over multiple object sets, requiring the model to reason about intersections, unions, complements, or cardinality comparisons. We design four subsets: (1) Joint-Set Aggregation: The question asks for the total count of objects from multiple categories (\textit{e.g.}, ``How many A and B together?''). The model must count each category separately and then sum, a simple but often overlooked compositional operation. (2) Complementary Exclusion: The model must count objects that belong to one set but not another. This tests the ability to apply logical negation in the visual domain. (3) Categorical Cardinality: The model counts the number of distinct object types or categories present in the image, independent of their instance frequencies. For example, ``How many kinds of fruits are in the image?'' This requires the model to recognize and categorize objects at a type level, then determine the cardinality of the set of categories, a higher-level cognitive skill beyond mere instance counting. (4) Differential Comparison: The model computes the numerical difference between two sets (\textit{e.g.}, ``How many more apples than oranges?''). This goes beyond comparison to exact arithmetic on perceived quantities. Existing benchmarks rarely include such logical compositions. When they do, they treat them as incidental rather than systematic. HoloCount provides controlled subsets for each operation, enabling fine-grained diagnosis of where models fail—whether in visual perception, logical parsing, or arithmetic execution. The building pipeline of these subsets is similar to that of semantic counting.

\subsubsection{Robustness Testing}
While Semantic Counting evaluates attribute-based filtering and Analytical Counting examines spatial and set-based logical reasoning, both operate under relatively clean, well-structured visual conditions. In real-world deployments, however, MLLMs face adverse perceptual domains (\textit{e.g.}, high density, occlusion, and aerial views) and deceptive priors (\textit{e.g.}, language biases, null targets, and visual illusions). To thoroughly assess the robustness of MLLMs' counting ability under such challenges, we construct seven additional subsets within HoloCount that explicitly target adverse visual conditions and counter-grounded linguistic priors.

\noindent \textbf{High-Density Enumeration.} For dense scenarios, we initially planned to sample images from existing high-density benchmarks (\textit{e.g.}, ShanghaiTech~\cite{zhang2016single} for crowd counting). However, we found that many of those samples are ambiguous and difficult to verify. Therefore, we developed a script to generate synthetic symbols on a blank canvas, mimicking dense object distributions. 

\noindent \textbf{Aerial Perspective Shift.} Most existing counting benchmarks focus on natural, ground-level viewpoints (\textit{e.g.}, street scenes, indoor environments). To evaluate counting under out-of-distribution (OOD) visual conditions, we sample images from CA44~\cite{jiang2024trex2}, where one subset provides aerial drone-view perspectives. These top-down views introduce significant changes in object appearance, scale, and spatial layout compared to typical counting datasets. The original data contain some noise and inaccurate QA pairs, so we manually filter out these problematic samples to ensure correctness.

\noindent \textbf{Small-Scale Enumeration.} Counting tiny objects is particularly challenging for MLLMs because such objects occupy few pixels, offering limited visual features for recognition and enumeration~\cite{Nikouei_2025}. Most existing benchmarks focus on reasonably sized objects, leaving small-scale counting under-evaluated. To address this gap, we sample images that contain objects smaller than 32$\times$32 pixels from the Object365~\cite{Object365} validation set. We then use an MLLM to filter out incorrect counting samples and employ human annotators to verify the final QA pairs.

\noindent \textbf{Partial Object Occlusion.} In real-world scenes, objects are often partially occluded, requiring models to infer hidden parts or count based on visible evidence~\cite{ouardirhi2024enhancing,yang2020learning}. However, current counting benchmarks rarely consider occlusion as a systematic variable. To evaluate counting under occlusion, we start from the validation set of Object365~\cite{Object365} and retain samples where objects exhibit heavy occlusion (IoU $>$ 0.85). We then use an MLLM to filter out incorrect counting samples and employ human annotators to verify the final QA pairs.

\noindent \textbf{Null-Target Prompting.} This subset evaluates whether a model hallucinates when asked about a concept absent from the image. In real-world applications, models may confidently produce a positive count for objects that do not exist, revealing a critical failure in visual grounding. To systematically probe this issue, we construct queries targeting categories that are either semantically related to the scene or completely out of context. Every query is carefully verified to have zero ground-truth instances. A robust model should correctly answer zero, whereas a hallucinating model may produce a non-zero response, exposing its reliance on language priors rather than genuine visual perception.

\noindent \textbf{Linguistic Prior Conflict.} In some cases, models may answer based on language priors rather than image content~\cite{lee2025vlindbenchmeasuringlanguagepriors,wu2025lanp,long2025understanding}. For example, a lobster typically has two claws. If we provide an image where a lobster has only one claw, does the model follow the visual evidence or the linguistic prior? Similarly, we probe common-sense knowledge such as the number of legs on a cat. This subset tests whether the model can override strong language priors when they conflict with the visual input. To construct this dataset, we first manually design a prompt set describing counterfactual objects (\eg, a lobster with one claw, a three-legged cat). We then use Gemini Image to generate corresponding images and build QA pairs, followed by human verification to ensure visual plausibility and correct ground-truth.

\noindent \textbf{Visual Distractor Illusion.} In this subset, objects are visually similar but represent different concepts, such as variations in material (plastic vs. real fruit). These subtle distinctions challenge the model's fine-grained discrimination ability, requiring it to count only the target category while ignoring visually distracting confounders. We collect samples from \cite{chen2026mvibenchcomprehensivebenchmarkevaluating}, then convert them to QA format with carefully designed instructions that specify the target attribute. For example, an image containing both real apples and plastic decorative apples may prompt the question: ``How many real apples are in this image?'' All QA pairs undergo rigorous manual verification to ensure unambiguous grounding.

\section{Experiments}
\label{sec:exp}
\subsection{Setup}
\noindent \textbf{Models.} We evaluate a comprehensive collection of state-of-the-art Multimodal Large Language Models (MLLMs) on our HoloCount benchmark. Our evaluation spans both open-source and proprietary frontier models to provide a holistic view of current counting capabilities. (1) Open-Source Models. We include widely adopted open-source MLLMs from leading families. Specifically, we evaluate the Qwen-VL series~\cite{bai2025qwen3-vl}, including Qwen2.5-VL~\cite{bai2025qwen25vltechnicalreport}, Qwen3-VL~\cite{bai2025qwen3-vl}, and Qwen3.5~\cite{qwen35blog}, InternVL series~\cite{wang2025internvl35advancingopensourcemultimodal}, Gemma3 series~\cite{gemmateam2025gemma3technicalreport}, MiniMax-M3, and Kimi series~\cite{kimiteam2026kimik25visualagentic}. All open-source models are evaluated with their official checkpoints under zero-shot settings without any task-specific fine-tuning. (2) Proprietary API Models. To benchmark against proprietary state-of-the-art systems, we evaluate leading commercial models via their official APIs. These include GPT-4.1, GPT-5.4, and GPT-5.5 from the GPT series~\cite{achiam2023gpt}, Gemini-2.5-Pro, Gemini-3-Flash-Preview and Gemini-3.1-Pro-Preview from the Gemini family, as well as the Claude family. For proprietary models, we use default generation parameters and follow each vendor's recommended best practices for visual question answering. 

\noindent \textbf{Evaluation Details.} For each model, we perform zero-shot inference on all 2,480 QA pairs across the 20 subsets of HoloCount. Models receive the input image and the counting question as a plain text prompt without any few-shot examples or chain-of-thought instructions; the system prompt can be seen in the Appendix. We extract the numerical answer from the model's generated response using regular expression matching. If the response contains no valid number, we record an incorrect prediction. Following prior work~\cite{countbench,tamarapalli2025countqa}, we report exact match accuracy as the percentage of samples where the predicted count exactly matches the ground-truth count. For open-source models, all experiments are conducted on NVIDIA H800 GPUs.

\subsection{Main Results and Analysis}

\begin{table}[tbp]
\centering
\caption{Main results on HoloCount benchmark. \textbf{Bold} and \underline{underline} denote the best and second-best results in each column, respectively.}
\label{tab:main_results}
\setlength{\tabcolsep}{2pt}
\renewcommand{\arraystretch}{1.2}
\resizebox{\textwidth}{!}{
\begin{tabular}{l|cccccc|ccccccc|ccccccc|c}
\toprule
& \multicolumn{6}{c|}{Semantic Counting} & \multicolumn{7}{c|}{Analytical Counting} & \multicolumn{7}{c|}{Robustness Testing} &  \\
Model & \rotatebox{80}{Atom.} & \rotatebox{80}{Chrom.} & \rotatebox{80}{Mater.} & \rotatebox{80}{Scale} & \rotatebox{80}{Act.$\&$St.} & \rotatebox{80}{Gen.} & \rotatebox{80}{Vis.} & \rotatebox{80}{Coord.} & \rotatebox{80}{Orient.} & \rotatebox{80}{Diff.} & \rotatebox{80}{Aggr.} & \rotatebox{80}{Excl.} & \rotatebox{80}{Categ.} & \rotatebox{80}{Dense} & \rotatebox{80}{Aerial} & \rotatebox{80}{Small} & \rotatebox{80}{Occl.} & \rotatebox{80}{Null.T.} & \rotatebox{80}{Ling.P.} & \rotatebox{80}{Vis.D.} & Avg. \\

\midrule
\rowcolor{gray!10}
\multicolumn{22}{l}{\textit{Open-Source}} \\
Gemma3-4B & 61.0 & 50.5 & 36.6 & 55.0 & 43.1 & 45.5 & 38.2 & 33.3 & 39.3 & 22.0 & 20.0 & 29.7 & 53.5 & 0.0 & 12.8 & 19.8 & 26.2 & 80.4 & 19.0 & 38.0 & 39.8 \\
Gemma3-12B & 64.8 & 52.0 & 59.4 & 64.0 & 61.8 & 45.5 & 35.3 & 37.2 & 51.2 & 37.0 & 44.0 & 32.7 & 58.4 & 1.0 & 11.0 & 22.8 & 32.0 & 72.8 & 23.3 & 45.5 & 45.0 \\
Gemma3-27B & 66.5 & 60.9 & 56.4 & 59.0 & 56.9 & 51.5 & 41.2 & 33.3 & 58.3 & 34.0 & 46.0 & 42.6 & 59.4 & 0.0 & 14.7 & 29.7 & 35.0 & 87.2 & 20.2 & 49.0 & 48.4 \\
Qwen2.5-VL-7B-Instruct & 76.4 & 57.9 & 61.4 & 70.0 & 66.7 & 65.3 & 50.0 & 25.6 & 65.5 & 35.0 & 31.0 & 46.5 & 65.3 & 2.0 & 18.3 & 29.7 & 36.9 & 95.6 & 20.9 & 60.5 & 52.9 \\
Qwen2.5-VL-32B-Instruct & 75.8 & 69.8 & 70.3 & 69.0 & 70.6 & 67.3 & 58.8 & 28.2 & 71.4 & 26.0 & 39.0 & 54.5 & 65.3 & 0.0 & 19.3 & 33.7 & 44.7 & 93.6 & 31.3 & 59.0 & 56.1 \\
Qwen2.5-VL-72B-Instruct & 80.8 & 72.3 & 74.3 & 81.0 & 75.5 & 64.4 & 63.7 & 14.1 & 67.9 & 51.0 & 61.0 & 60.4 & 67.3 & 1.0 & 23.9 & 43.6 & 58.3 & 95.2 & 36.8 & 67.0 & 61.6 \\
InternVL3.5-8B & 74.2 & 61.9 & 67.3 & 70.0 & 64.7 & 60.4 & 40.2 & 44.9 & 63.1 & 37.0 & 27.0 & 51.5 & 59.4 & 4.0 & 13.8 & 30.7 & 32.0 & 92.8 & 21.5 & 62.5 & 52.6 \\
InternVL3.5-38B & 79.7 & 66.8 & 74.3 & 72.0 & 75.5 & 66.3 & 68.6 & 61.5 & 78.6 & 35.0 & 44.0 & 59.4 & 73.3 & 1.0 & 28.4 & 40.6 & 54.4 & 96.0 & 27.0 & 61.0 & 60.6 \\
Qwen3-VL-8B-Instruct & 78.0 & 65.8 & 70.3 & 73.0 & 78.4 & 63.4 & 57.8 & 66.7 & 65.5 & 29.0 & 33.0 & 49.5 & 67.3 & 0.0 & 22.0 & 32.7 & 43.7 & 96.4 & 37.4 & 64.0 & 58.1 \\
Qwen3-VL-32B-Instruct & 84.1 & 72.8 & 70.3 & 77.0 & 82.4 & 65.3 & 65.7 & 66.7 & 72.6 & 37.0 & 44.0 & 60.4 & 62.4 & 3.0 & 26.6 & 33.7 & 47.6 & \textbf{98.0} & 42.3 & 71.0 & 62.7 \\
Kimi-K2.5 & 89.0 & 78.2 & 75.2 & 82.0 & 80.4 & 68.3 & 57.8 & 71.8 & 76.2 & 66.0 & 67.0 & 63.4 & 82.2 & 3.0 & 67.0 & 37.6 & 46.6 & 94.4 & 44.8 & 73.5 & 68.8 \\
Kimi-K2.6 & 89.0 & 83.7 & 83.2 & 84.0 & 80.4 & 82.2 & 67.6 & 67.9 & 76.2 & 70.0 & 68.0 & 74.3 & \textbf{89.1} & 4.0 & 68.8 & 40.6 & 61.2 & 95.6 & 48.5 & \underline{78.5} & 73.0 \\
MiniMax-M3 & 74.2 & 70.8 & 64.4 & 66.0 & 67.6 & 66.3 & 52.0 & 44.9 & 57.1 & 56.0 & 57.0 & 51.5 & 75.2 & 1.0 & 41.3 & 30.7 & 39.8 & 89.6 & 35.6 & 69.0 & 58.9 \\
Qwen3.5-4B & 80.8 & 78.7 & 77.2 & 78.0 & 85.3 & 76.2 & 66.7 & 59.0 & 71.4 & 67.0 & 71.0 & 70.3 & 80.2 & 2.0 & 71.6 & 42.6 & 55.3 & 87.6 & 44.2 & 70.5 & 68.6 \\
Qwen3.5-9B & 86.8 & 83.2 & 80.2 & \underline{85.0} & 83.3 & 75.2 & 74.5 & 64.1 & 76.2 & 67.0 & 69.0 & 67.3 & 81.2 & 3.0 & 77.1 & 50.5 & \underline{64.1} & 86.0 & 42.3 & 76.5 & 71.4 \\
Qwen3.5-27B & 90.7 & 86.1 & \textbf{85.1} & \textbf{86.0} & 87.3 & 83.2 & 71.6 & 80.8 & \textbf{82.1} & 76.0 & 79.0 & 76.2 & 80.2 & \underline{20.0} & \textbf{82.6} & 52.5 & 61.2 & 89.6 & 52.1 & 73.0 & \underline{75.9} \\
Qwen3.5-35B-A3B & 90.7 & 87.1 & 83.2 & 83.0 & 84.3 & 83.2 & 71.6 & 78.2 & 78.6 & 69.0 & 77.0 & 78.2 & 82.2 & 5.0 & 78.0 & 60.4 & \textbf{65.0} & 90.4 & 46.6 & 77.5 & 75.0 \\
Qwen3.5-122B-A10B & 90.7 & 86.1 & 82.2 & 84.0 & \textbf{89.2} & 77.2 & 75.5 & 74.4 & 77.4 & 70.0 & \textbf{82.0} & 79.2 & 82.2 & 9.0 & 78.9 & 58.4 & \underline{64.1} & 90.8 & 47.9 & 78.0 & 75.4 \\
Qwen3.5-397B-A17B & 92.9 & \textbf{90.1} & 83.2 & 82.0 & \textbf{89.2} & 80.2 & 76.5 & \underline{83.3} & \textbf{82.1} & 74.0 & 78.0 & \underline{80.2} & 80.2 & 14.0 & 78.9 & 60.4 & 61.2 & 90.8 & 50.3 & \textbf{80.0} & \textbf{76.9} \\
\midrule
\rowcolor{gray!10}
\multicolumn{22}{l}{\textit{Closed-Source}} \\
Claude Sonnet 4.5 & 68.1 & 58.9 & 57.4 & 61.0 & 58.8 & 58.4 & 40.2 & 16.7 & 45.2 & 39.0 & 41.0 & 42.6 & 64.4 & 2.0 & 20.2 & 33.7 & 34.0 & \underline{97.2} & 35.0 & 57.5 & 51.2 \\
Claude Sonnet 4.6 & 74.2 & 67.3 & 63.4 & 66.0 & 75.5 & 58.4 & 51.0 & 42.3 & 67.9 & 34.0 & 42.0 & 47.5 & 71.3 & 1.0 & 24.8 & 27.7 & 44.7 & 95.2 & 36.8 & 65.0 & 56.7 \\
Claude Opus 4.6 & 72.5 & 59.9 & 70.3 & 68.0 & 66.7 & 61.4 & 40.2 & 39.7 & 60.7 & 52.0 & 41.0 & 53.5 & 66.3 & 0.0 & 13.8 & 29.7 & 45.6 & 94.4 & 41.1 & 60.0 & 55.4 \\
Claude Opus 4.7 & 76.9 & 67.3 & 72.3 & 73.0 & 77.5 & 63.4 & 56.9 & 38.5 & 69.0 & 58.0 & 62.0 & 53.5 & \underline{84.2} & 4.0 & 27.5 & 40.6 & 50.5 & 95.6 & 49.1 & 70.5 & 62.8 \\
Claude Opus 4.8 & 82.4 & 72.3 & 75.2 & 77.0 & 71.6 & 68.3 & 48.0 & 39.7 & 65.5 & 59.0 & 58.0 & 63.4 & 82.2 & 1.0 & 7.3 & 38.6 & 55.3 & 90.4 & 52.1 & 73.5 & 62.6 \\
GPT-4.1 & 77.5 & 71.3 & 67.3 & 73.0 & 70.6 & 61.4 & 61.8 & 42.3 & 59.5 & 49.0 & 57.0 & 60.4 & 77.2 & 2.0 & 26.6 & 23.8 & 50.5 & 88.0 & 46.6 & 73.0 & 60.5 \\
GPT-5.4 & 73.1 & 67.3 & 70.3 & 65.0 & 76.5 & 63.4 & 57.8 & 38.5 & 65.5 & 55.0 & 55.0 & 61.4 & 57.4 & 8.0 & 39.4 & 30.7 & 44.7 & 85.6 & 46.6 & 64.0 & 59.2 \\
GPT-5.5 & 84.6 & 73.3 & 76.2 & 73.0 & 84.3 & 68.3 & 59.8 & 69.2 & 73.8 & 58.0 & 60.0 & 66.3 & 81.2 & \underline{20.0} & 58.7 & 42.6 & 43.7 & 88.8 & 48.5 & 76.0 & 67.6 \\

Gemini-2.5-Pro & 84.1 & 74.3 & 76.2 & 74.0 & 81.4 & 71.3 & 71.6 & 56.4 & 66.7 & 65.0 & 65.0 & 74.3 & 82.2 & 0.0 & 63.3 & 39.6 & 56.3 & 93.4 & 44.2 & 71.0 & 67.9 \\
Gemini-3-Flash-Preview & \underline{93.4} & \textbf{90.1} & 77.2 & \underline{85.0} & 82.4 & \underline{86.1} & \underline{77.5} & \textbf{84.6} & 78.6 & \underline{78.0} & 77.0 & \underline{80.2} & 81.2 & 12.0 & 78.9 & \textbf{74.3} & 61.2 & 62.4 & \underline{69.9} & 67.5 & 74.8 \\
Gemini-3.1-Pro-Preview & \textbf{95.1} & 89.1 & \underline{84.2} & \underline{85.0} & 85.3 & \textbf{87.1} & \textbf{86.3} & 9.0 & 81.0 & \textbf{79.0} & \textbf{82.0} & \textbf{84.2} & 74.3 & \textbf{49.0} & \textbf{82.6} & \underline{73.3} & 63.1 & 55.2 & \textbf{73.6} & 67.0 & 74.7 \\
\bottomrule
\end{tabular}
}

\end{table}

\noindent \textbf{Overall Performance Comparison.} Table~\ref{tab:main_results} reveals that the most advanced open-source models now rival or even surpass proprietary systems on the HoloCount benchmark. The top open-source entry, Qwen3.5-397B-A17B, achieves a macro average of 76.9\%, closely followed by Qwen3.5-27B (75.9\%) and Qwen3.5-122B-A10B (75.4\%). Among closed-source models, Gemini-3-Flash-Preview and Gemini-3.1-Pro-Preview lead with 74.8\% and 74.7\%, respectively. This near tie indicates that the performance gap between well-trained open-source families and commercial APIs has substantially narrowed. Scaling remains effective: within the Qwen3.5 family, moving from 4B to 27B yields a gain of $7.3\%$, and the largest 397B MoE model further pushes the state-of-the-art.

\begin{table}[h]
\centering
\caption{Performance on the high-density counting subset. We report Accuracy (exact match), Mean Absolute Error (MAE), and Relative Accuracy at 5\% and 10\% thresholds (RA@5\%, RA@10\%).}
\label{tab:high_density}
\resizebox{0.7\textwidth}{!}{
\begin{tabular}{lcccc}
\toprule
\textbf{Model} & \textbf{Acc (\%)} $\uparrow$ & \textbf{MAE} $\downarrow$ & \textbf{RA@5\% (\%)} $\uparrow$ & \textbf{RA@10\% (\%)} $\uparrow$ \\
\midrule
Qwen3-VL-8B-Instruct & 0.0 & 40.11 & 10.0 & 18.0 \\
Qwen3-VL-32B-Instruct & 3.0 & 32.94 & 19.0 & 32.0 \\
Qwen2.5-VL-72B-Instruct & 1.0 & 38.66 & 5.0 & 12.0 \\
Gemma3-27B & 0.0 & 50.08 & 2.0 & 6.0 \\
Gemini-2.5-Pro & 0.0 & 32.56 & 6.0 & 15.0 \\
Kimi-K2.5 & 3.0 & 32.59 & 13.8 & 28.7 \\
Gemini-3.1-Pro-Preview & \textbf{49.0} & \textbf{10.46} & \textbf{79.0} & \textbf{82.0} \\
\bottomrule
\end{tabular}
}
\end{table}

\noindent \textbf{The Catastrophic Failure Mode of High-Density Contexts.} Counting in high-density scenes (\texttt{Dense}) remains a universal weakness. Even the strongest models show dramatic degradation. For example, Qwen3.5-27B achieves only 20.0\% exact accuracy on this subset, while its average across all other subsets exceeds 80\%. Most open-source models score below 10\%, and several (\eg, Gemma3-27B, Qwen2.5-VL-72B) fall to 0--1\%. The best closed-source performer, Gemini-3.1-Pro-Preview, reaches 49.0\% but is still far from reliable. To further diagnose this breakdown, Table~\ref{tab:high_density} reports Mean Absolute Error (MAE) and Relative Accuracy at 5\% and 10\% thresholds. Open-source models exhibit MAE values of 32--50, meaning their predictions deviate by roughly one-third to half of the true count. RA@5\% stays below 20\% for all but Gemini-3.1-Pro-Preview, which achieves 79.0\% RA@5\% and a low MAE of 10.46. These results confirm that dense scenes cause systematic under-counting, likely because current vision encoders cannot maintain fine-grained spatial distinctions when objects heavily overlap or crowd the canvas.

\noindent \textbf{Logical Execution vs. Baseline Perception Disconnect.} A clear dissociation exists between atomic counting and higher-order analytical tasks. Nearly all models perform strongly on Atomic Counting (\texttt{Atom.}), often exceeding 85\%. However, performance drops sharply on tasks that demand set operations or comparisons. For instance, InternVL3.5-38B falls from 79.7\% on atomic counting to 35.0\% on Differential Comparison (\texttt{Diff.}); Qwen3.5-27B drops from 90.7\% to 76.0\%. Even the best models show a consistent gap, confirming that the primary bottleneck is not visual perception per se but the symbolic reasoning and arithmetic composition required after objects are perceived.

\noindent \textbf{The Inverse Performance Paradox in Null-Target Presentation.} The \texttt{Null.T.} subset reveals a striking inversion: smaller or older open-source models often outperform advanced proprietary engines when asked to count objects that are entirely absent. Lightweight models such as Qwen3-VL-8B-Instruct (96.4\%) and Qwen2.5-VL-7B-Instruct (95.6\%) achieve near-perfect scores. In contrast, Gemini-3-Flash-Preview drops to 62.4\%, and Gemini-3.1-Pro-Preview collapses to 55.2\%---the lowest among all models. This paradox suggests that aggressive optimization for detailed scene description and object retrieval inadvertently increases over-confidence in the presence of any visual signal. Larger models become reluctant to output zero, even when no target exists, whereas simpler models faithfully follow the instruction.

\noindent \textbf{Resisting the Siren Call of Linguistic Priors.} The Linguistic Prior Conflict (\texttt{Ling.P.}) subset tests whether models can override strong language priors (\textit{e.g.}, ``a lobster has two claws'') when the image contradicts them. Most models perform poorly, with scores in the 20--40\% range for Gemma3, Qwen2.5, and early Qwen3 families. This indicates a general ``lazy'' reliance on textual statistics rather than visual verification. However, the Gemini-3 series stands out: Gemini-3-Flash-Preview achieves 69.9\%, and Gemini-3.1-Pro-Preview achieves 73.6\%, the highest among all evaluated models. This gap implies that the Gemini architecture incorporates a stronger vision-first arbitration mechanism that can suppress incompatible linguistic priors when they conflict with observed pixels.

\noindent \textbf{Vulnerability to Visual Distractors and False-Positive Susceptibility.} The Visual Distractor Illusion (\texttt{Vis.D.}) subset measures resilience against visually similar but semantically distinct distractors (\eg, real vs. plastic fruit). All models suffer a significant drop compared to atomic counting. Smaller models are hit hardest: Gemma3-4B scores only 38.0\%. Among open-source models, the best performer is Qwen3.5-397B-A17B (80.0\%), while leading closed-source models like Gemini-3-Flash-Preview (67.5\%) and GPT-5.5 (76.0\%) also show substantial room for improvement. This pattern indicates that although modern vision encoders capture global scene semantics, their attention mechanisms often fail to perform fine-grained attribute filtering, leading to over-counting of distractors that share a coarse visual match.

\begin{figure}
    \centering
    \includegraphics[width=0.7\linewidth]{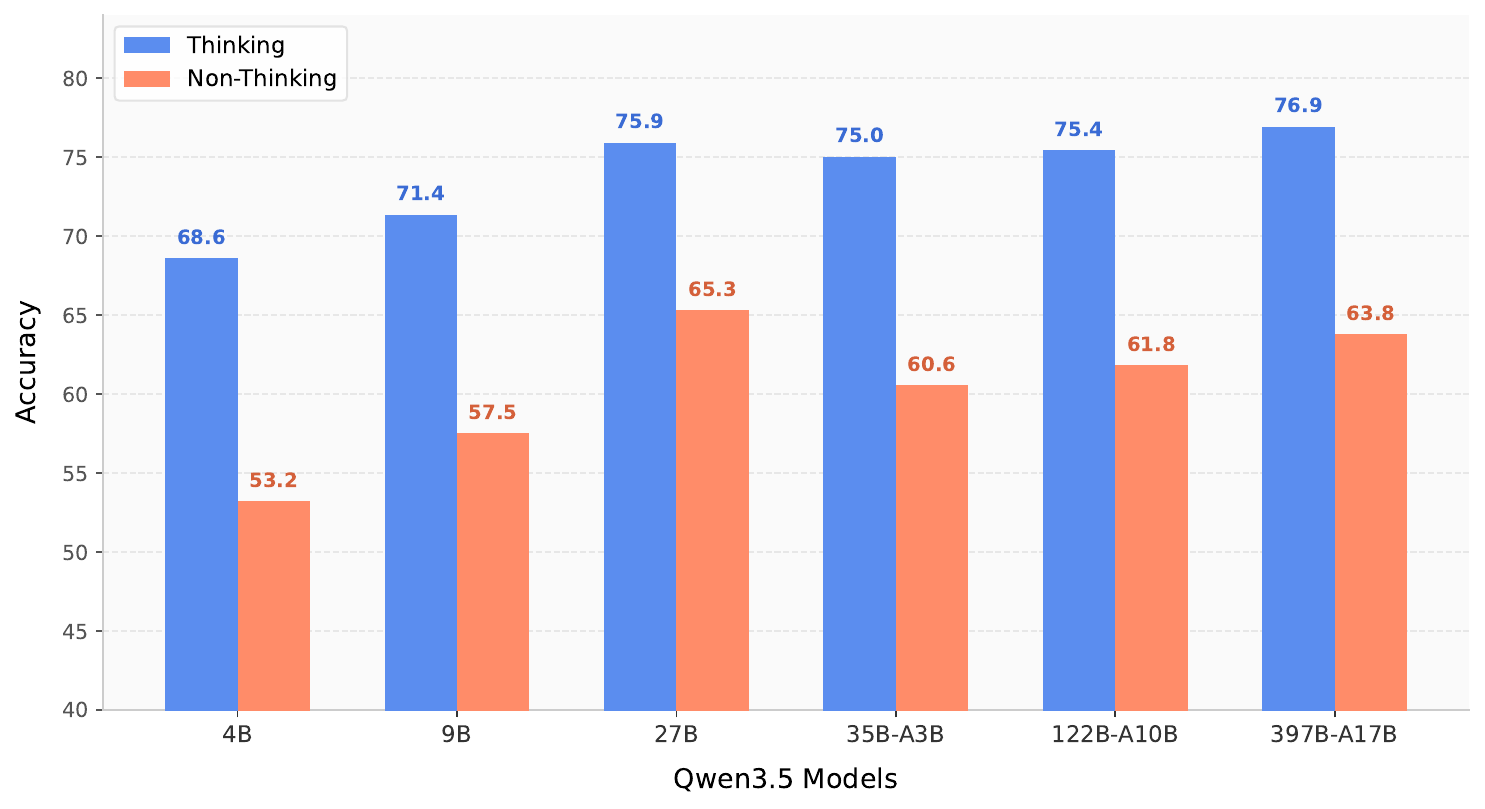}
    \caption{Comparison of Qwen3.5 models in thinking versus non-thinking (instruct) mode on HoloCount. Thinking mode consistently improves counting accuracy across all model scales.} 
    \label{fig:thinking}
\end{figure}

\noindent \textbf{Impact of Thinking Mode.} We investigate the effect of extended thinking on counting performance by comparing Qwen3.5 models in thinking mode versus non-thinking (instruct) mode across six model scales. As shown in Fig.~\ref{fig:thinking}, thinking mode consistently improves counting accuracy across all scales, with gains ranging from 10.6 to 15.4 absolute percentage points. Notably, smaller models benefit more (4B: +15.4, 9B: +13.9), but even the largest 397B model gains +13.1 points. This suggests that counting tasks require deliberate step-by-step reasoning, \eg, enumerating objects, filtering by attributes, or performing set operations, which is poorly approximated by direct pattern matching. Thinking mode effectively compensates for the lack of built-in counting primitives in standard MLLMs.

\section{Conclusion}
\label{sec:conclusion}

Existing counting benchmarks suffer from flat structures, narrow visual concepts, and a focus on perceptual enumeration, leaving complex failures largely undiagnosed. To bridge this gap, we introduced HoloCount, a hierarchical benchmark organized into three core pillars: Semantic Counting, Analytical Counting, and Robustness Testing. With 20 fine-grained subsets and 2,480 carefully curated QA pairs spanning 1,481 visual concepts, HoloCount provides a decoupled evaluation framework that isolates where and why models fail. Our extensive evaluation of over 20 MLLMs confirms that current models remain far from reliable counting, underscoring the urgent need for architectures that prioritize cognitive reasoning and resilience over superficial perception.





{\small
\bibliographystyle{plain}
\bibliography{egbib.bib}
}

\newpage
\appendix

\section{Benchmark Statistics}
In this section, we conduct a thorough statistical analysis of our benchmark to provide a holistic understanding of its composition and inherent properties. A comprehensive examination of these statistics is indispensable, as it not only reveals the diversity and potential biases of the dataset but also establishes a foundational context for interpreting the experimental results in subsequent sections. Our analysis is structured along three key perspectives: the overall count distribution, the semantic coverage of visual concepts, and the numerical scale range for Small-Scale Enumeration tasks. 

\subsection{Count Distribution}

\begin{figure}[h]
    \centering
    \includegraphics[width=0.8\linewidth]{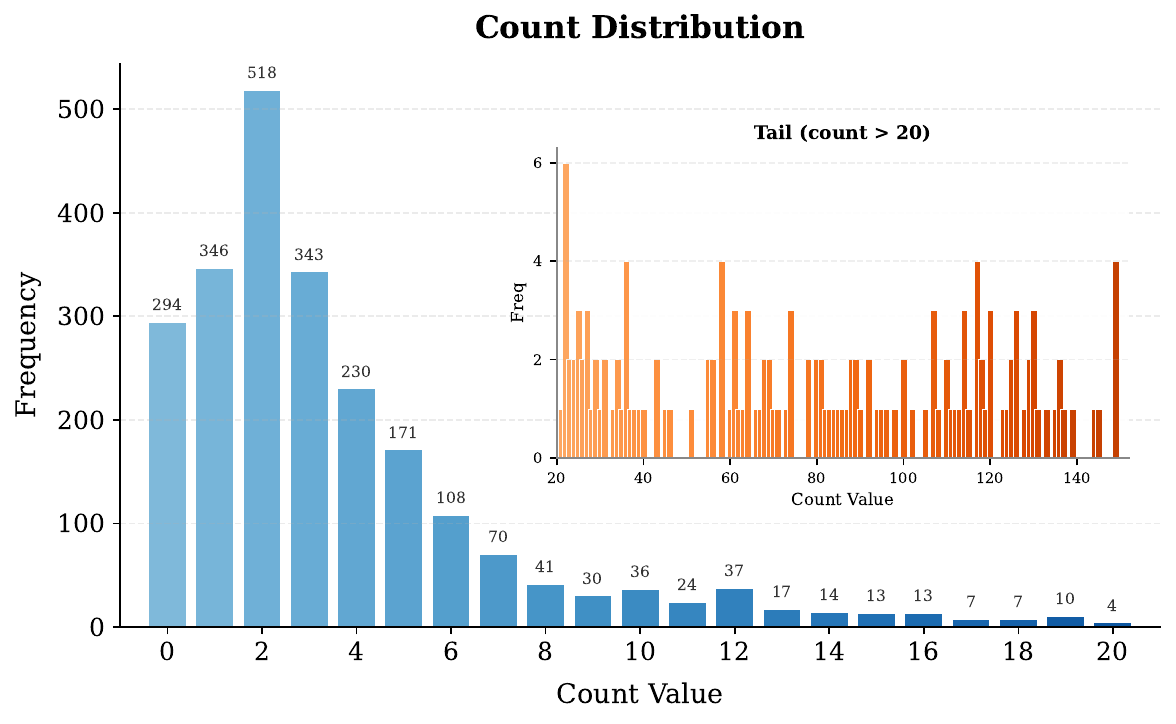}
    \caption{Distribution of ground-truth count values in HoloCount. The main plot shows the frequency of counts in [0, 20], while the inset displays the long-tail distribution for      
  counts beyond 20.} 
    \label{fig:count_dist}
\end{figure}

Fig.~\ref{fig:count_dist} illustrates the distribution of ground-truth count values across all 2,480 samples in HoloCount. The counts span a wide range from 0 to 149, with a mean of 8.06 and a median of 3. The distribution exhibits a long-tailed pattern: 64.8\% of samples fall within the [1, 5] range, reflecting common real-world counting scenarios, while 11.9\% of samples have a count of 0, mainly corresponding to the Null-Target Presentation subset where queried objects are absent from the image. Beyond 20, the tail extends to 149, primarily contributed by the dense and aerial counting subsets. This deliberate design covers both low-count fine-grained discrimination and high-count estimation, enabling comprehensive evaluation of counting capabilities across diverse difficulty levels.

\subsection{Visual Concept Word Cloud}
\begin{figure}[h]
    \centering
    \includegraphics[width=0.8\linewidth]{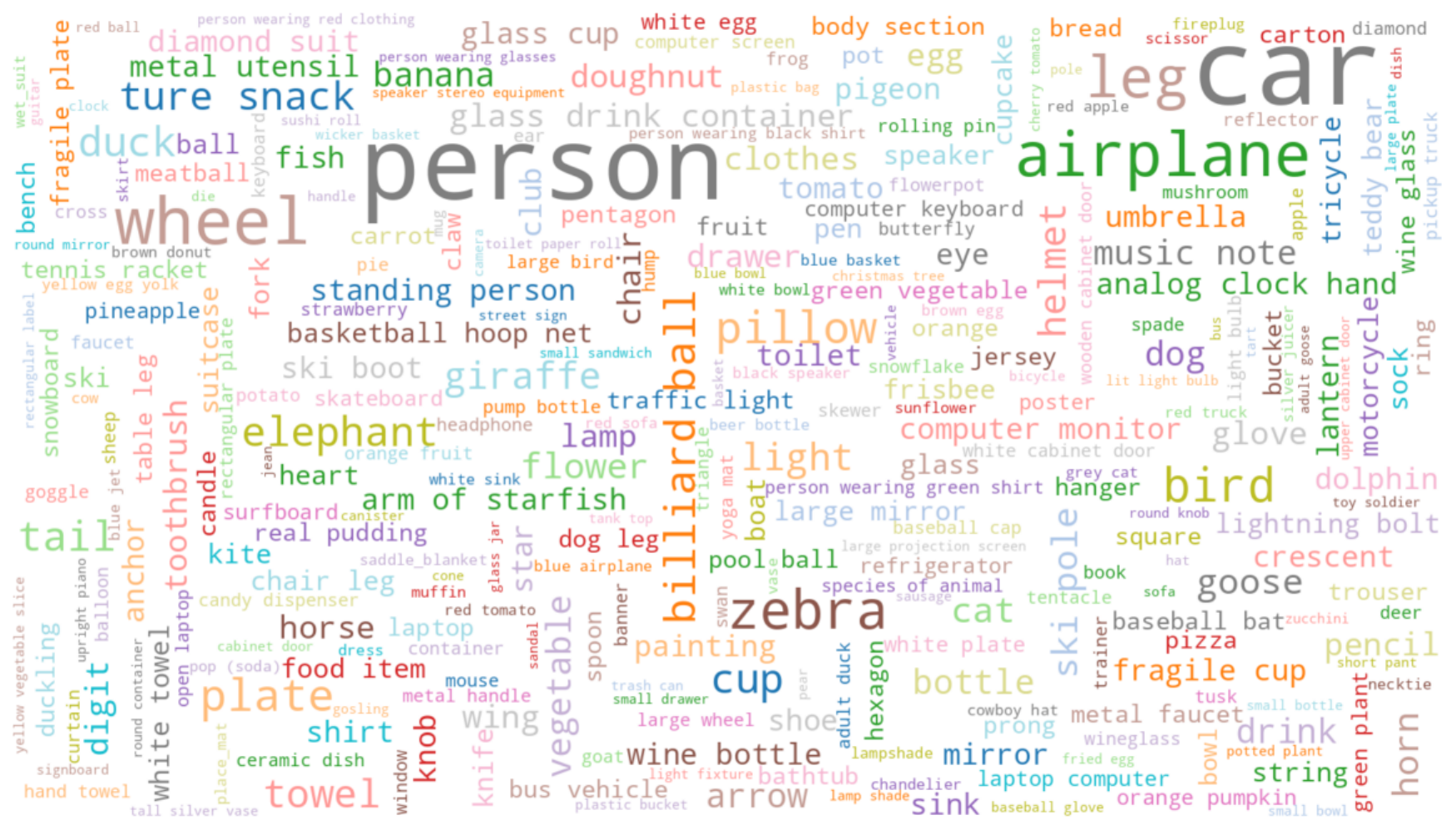}
    \caption{Word cloud of visual concepts in HoloCount. The benchmark covers 1,481 unique visual concepts spanning objects, animals, body parts, and fine-grained attributes.}
    \label{fig:wordcloud}
\end{figure}

\noindent \textbf{Visual Concept Extraction.} To quantify the semantic diversity of HoloCount, we extract the core visual concepts from each counting question using GPT-3.5-turbo (see Fig.~\ref{fig:prompt_concept_extraction}). Specifically, we prompt the model with a system instruction and few-shot examples to parse noun phrases from questions while normalizing to singular form, preserving distinguishing modifiers (\eg, ``yellow flower'', ``wooden chair''), and removing spatial or locational phrases (\eg, ``on the wall'', ``in the foreground''). For composite questions involving comparison or aggregation (\eg., ``How many more X than Y?''), all compared concepts are extracted independently. This process yields 1,481 unique visual concepts in 2,480 samples. 

\textbf{ Visual Concept Distribution.} Fig.~\ref{fig:wordcloud} visualizes the extracted concepts as a word cloud. The distribution reveals broad semantic coverage spanning common objects (car, person, cup, plate), animals (zebra, bird, elephant, duck, cat), body parts (leg, tail, wheel), and fine-grained attributes. Notably, the distribution is highly long-tailed: 74.7\% of concepts (1,105 out of 1,481) appear only once, indicating that HoloCount avoids over-concentration on a few dominant categories and provides broad coverage of the visual world.

\begin{figure}[t]
\begin{tcolorbox}[colback=gray!5, colframe=gray!50, title=System Prompt, fonttitle=\bfseries\small]
\small
You are an expert at extracting visual concepts from counting questions. Given a counting question, extract the core visual concept(s) that need to be counted. Rules:
\begin{enumerate}[leftmargin=*, nosep]
    \item Return ONLY the visual concept noun phrases, separated by commas if multiple.
    \item Use singular form (e.g., ``chair'' not ``chairs'', ``tile'' not ``tiles'').
    \item Include modifiers that distinguish the concept (e.g., ``yellow flower'', ``action card'', ``wooden chair'') but remove spatial/location phrases (e.g., ``on the wall'', ``in the foreground'').
    \item For comparison questions (e.g., ``How many more X than Y?''), extract all compared concepts.
    \item Do NOT include articles (a, an, the) or quantities.
    \item Output ONLY the concept(s), nothing else.
\end{enumerate}
\end{tcolorbox}

\vspace{4pt}
\begin{tcolorbox}[colback=blue!3, colframe=blue!30, title=Few-shot Examples, fonttitle=\bfseries\small]
\small
\textbf{Q:} How many tiles are on the wall with the shower? \\
\textbf{A:} tile \\[4pt]
\textbf{Q:} How many chairs and tables are there? \\
\textbf{A:} chair, table \\[4pt]
\textbf{Q:} How many more action cards are there than cat cards? \\
\textbf{A:} action card, cat card \\[4pt]
\textbf{Q:} How many yellow flowers are there? \\
\textbf{A:} yellow flower \\[4pt]
\textbf{Q:} Ignore the tubs in the back. How many tubs are there in the foreground? \\
\textbf{A:} tub
\end{tcolorbox}
\caption{Prompt used for visual concept extraction via GPT-3.5-turbo. The system prompt instructs the model to parse core visual noun phrases from counting questions, normalize to singular form, and preserve distinguishing modifiers while removing spatial references.}
\label{fig:prompt_concept_extraction}
\end{figure}

\subsection{Scale Distribution for Small-Scale Enumeration}
\begin{figure}[h]
    \centering
    \includegraphics[width=0.8\linewidth]{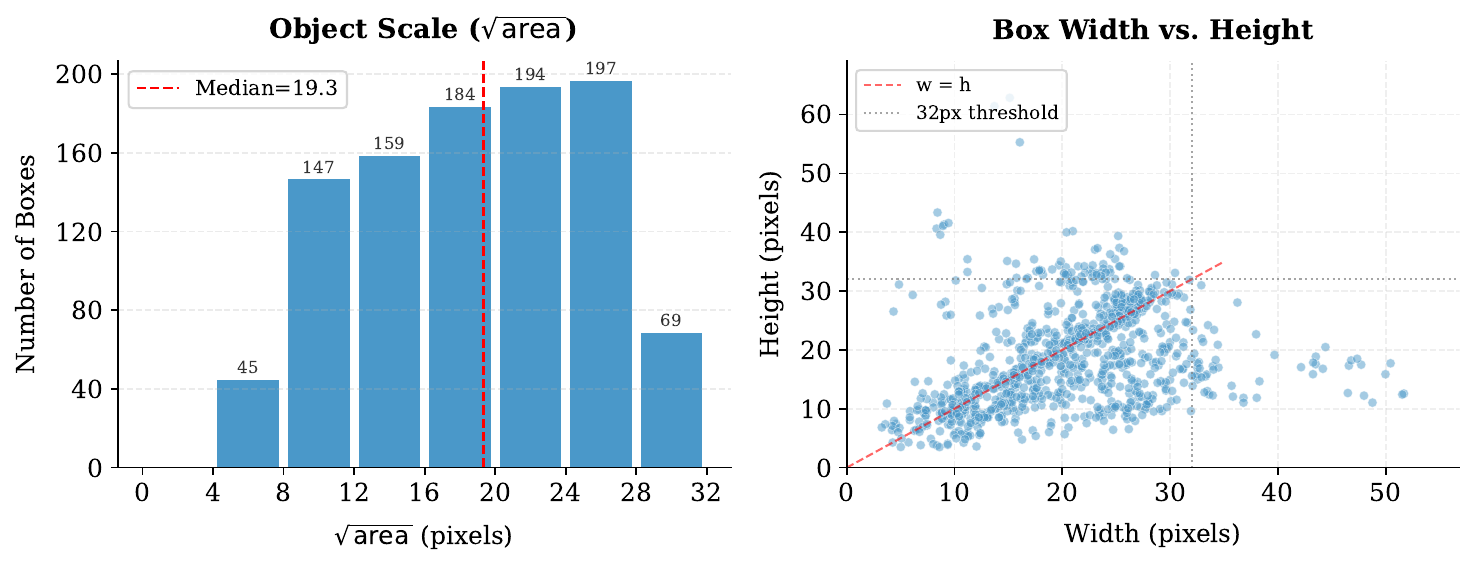}
    \caption{Object scale distribution in the small-scale enumeration subset (101 samples, 995 boxes). Left: histogram of $\sqrt{\text{area}}$ in pixels (median=19.3px, all $\leq$32px). Right: scatter plot of box width vs. height, with the 32px threshold indicated by dashed lines.}
    \label{fig:small_scale_dist}
\end{figure}

Fig.~\ref{fig:small_scale_dist} shows the object scale distribution in the small-scale enumeration subset (101 samples, 995 bounding boxes), where all targets are constrained to an area of $\leq$1024 pixels$^2$ (equivalent to 32$\times$32). The $\sqrt{\text{area}}$ ranges from 4.2 to 31.9 pixels with a median of 19.3 pixels. The distribution is concentrated between 8--32 pixels: 30.8\% of objects fall in the [8, 16) pixel range, 38.0\% in [16, 24), and 26.7\% in [24, 32), while 4.5\% are extremely small ($<$8 pixels). This subset specifically targets the challenging regime where objects occupy less than 0.16\% of the image area on average, testing whether models can detect and enumerate targets that are barely perceptible at typical display resolutions.

\section{Data Construction Details}
\subsection{QA Generation Prompts}
\label{sec:qa_gen_prompts}

Unlike basic object counting, which can be derived from existing detection datasets with bounding box annotations, attribute-based semantic counting and analytic counting lack readily available datasets. To construct these subsets, we leverage Gemini-3-Flash-Preview to generate question-answer pairs. This LLM-assisted generation paradigm enables rapid and scalable construction of counting datasets: by simply modifying the prompt templates, one can efficiently produce large-scale QA pairs covering diverse counting dimensions without labor-intensive manual annotation. The prompts used for generation are shown in Fig.~\ref{fig:prompt_attr_qa} and Fig.~\ref{fig:prompt_reasoning_qa}. Note that the sub-capability names in the prompts may differ slightly from those used in the final benchmark taxonomy, as the prompts were designed to guide QA generation rather than to define the evaluation taxonomy.

\begin{figure*}[t]
\begin{tcolorbox}[colback=gray!5, colframe=gray!50, title=Prompt for Attribute-based Counting QA Generation, fonttitle=\bfseries\small]
\small
You are an expert at creating counting QA pairs for evaluating vision-language models. Based on this image, generate 5 \textbf{attribute-based counting} question-answer pairs. Requirements:

\begin{enumerate}[leftmargin=*, nosep]
    \item \textbf{Capability}: All questions must be \textbf{conditional\_counting} with an \textit{attribute} constraint (what to count), not spatial (where to count). Use exactly one attribute type per question. Attribute sub\_capability and priority when multiple apply:
    \begin{itemize}[nosep]
        \item \textbf{color}: object color, hair/skin/appearance (e.g. ``How many red cars?'', ``How many people with blonde hair?''). Use for any color or appearance-related attribute.
        \item \textbf{state\_or\_action}: pose, action, state (e.g. ``How many people standing?'', ``How many dogs running?'').
        \item \textbf{size}: physical size only --- big, small, large (e.g. ``How many large balls?''). Do NOT use size for color/appearance.
        \item \textbf{material}: e.g. ``How many wooden chairs?''
        \item \textbf{other\_attribute}: any other attribute that does not fit above (e.g. pattern, shape).
    \end{itemize}

    \item \textbf{Format}: For each QA pair output one line in this exact format (no extra text): \\
    \texttt{\{``question'': ``<English counting question>'', ``count'': <integer>, ``sub\_capability'': ``<one of: color, size, material, state\_or\_action, other\_attribute>''\}}

    \item \textbf{Content}: Questions must be answerable by counting visible objects in the image that satisfy the attribute. Count must be correct for this image. Vary the attribute types across the list (include at least color and one other type if the image allows).
\end{enumerate}
\end{tcolorbox}
\caption{Prompt used for generating attribute-based conditional counting QA pairs. The model receives this system prompt along with an input image and produces structured QA outputs.}
\label{fig:prompt_attr_qa}
\end{figure*}

\begin{figure*}[t]
\begin{tcolorbox}[colback=gray!5, colframe=gray!50, title=Prompt for Reasoning-based Counting QA Generation, fonttitle=\bfseries\small]
\small
You are an expert at creating \textbf{reasoning-based counting} QA pairs for evaluating vision-language models. Based on this image, generate 3 question-answer pairs that require comparison, aggregation, exclusion, or other reasoning beyond simple direct counting.

\vspace{4pt}
\textbf{Capability \& Sub-dimensions} (use exactly one per question):

\begin{enumerate}[leftmargin=*, nosep]
    \item \textbf{comparison}: Compare quantities between two or more categories (who has more/fewer, or by how many). \\
    \textit{Examples}: ``How many more red cars than blue cars?'', ``Which color has the most balls?''

    \item \textbf{category\_kind}: Ask ``how many kinds/types'' (count of distinct categories, NOT instance count). \\
    \textit{Examples}: ``How many different fruits are there?'', ``How many types of vehicles appear?''

    \item \textbf{part\_whole}: Part vs whole (complete vs slices, pieces, or segments). \\
    \textit{Examples}: ``How many whole pizzas are there?'', ``How many slices?''

    \item \textbf{aggregation}: Multi-category or multi-condition count, then sum. \\
    \textit{Examples}: ``How many people in red and blue shirts in total?'', ``What is the total number of dogs and cats?''

    \item \textbf{exclusion}: Count from the full set after excluding a specific condition. \\
    \textit{Examples}: ``How many balls are there besides the blue ones?'', ``Excluding the smallest, how many apples remain?''

    \item \textbf{other\_reasoning}: Other complex reasoning: multi-step analysis, joint attributes, cross-region counting. \\
    \textit{Examples}: ``How many objects are both red and large?''
\end{enumerate}

\vspace{4pt}
\textbf{Response Format}: Output one JSON object per line: \\
\texttt{\{``question'': ``<English counting question>'', ``count'': ``<number or short text>'', ``sub\_capability'': ``<one of: comparison, category\_kind, part\_whole, aggregation, exclusion, other\_reasoning>''\}}

\vspace{4pt}
\textbf{Requirements}:
\begin{enumerate}[leftmargin=*, nosep]
    \item All questions must be answerable by analyzing visible objects in the image. No external knowledge.
    \item Cover at least 3 different sub\_capabilities across the set; vary question types.
    \item Ensure counts and comparisons are correct for this specific image.
    \item Avoid trivial direct-count questions. Focus on questions that require comparison, aggregation, exclusion, or multi-step reasoning.
\end{enumerate}
\end{tcolorbox}
\caption{Prompt used for generating reasoning-based counting QA pairs. The model is instructed to produce diverse question types covering comparison, aggregation, exclusion, and category-level reasoning.}
\label{fig:prompt_reasoning_qa}
\end{figure*}

\subsection{Linguistic Prior Conflict Image Generation Pipeline}
In some cases, models may answer based on language priors rather than image content. For example, a lobster typically has two claws --- if we provide an image where a lobster has only one claw, does the model follow the visual evidence or the linguistic prior? To systematically evaluate this behavior, we construct a linguistic prior conflict subset through the following pipeline. We first manually curate a triplet (object, part, prior\_count) encoding strong common-sense associations (\eg, dog/legs/4, spider/legs/8). For each triplet, we generate abnormal counts that deviate from the prior (typically $\pm$1 to $\pm$3), then fill them into prompt templates that explicitly request photorealistic images depicting the specified abnormal quantity with all parts fully visible. We use Gemini to generate images from these prompts, followed by manual verification to retain only images where the ground-truth count is visually unambiguous. The resulting evaluation question takes the form ``How many \{part\} does the \{object\} in the image have?'', a model relying on linguistic priors would answer with the prior count, while a faithful visual perceiver would report the actual depicted count.

\subsection{Counting System Prompt}
During evaluation, all models receive the same system prompt to ensure a fair and consistent output format across different architectures. The prompt is designed to elicit a single integer response, minimizing ambiguity in answer extraction:

\begin{tcolorbox}[colback=gray!5, colframe=gray!50, title=Evaluation System Prompt, fonttitle=\bfseries\small]
\small
You are a helpful assistant that counts the number of items in an image. The user will provide an image and ask a question about the number of a certain type of item in the image. Your output should STRICTLY be a single integer and nothing else.
\end{tcolorbox}

This strict integer-only constraint avoids the need for post-hoc answer parsing (e.g., extracting numbers from natural language responses) and ensures that evaluation metrics directly reflect the model's counting ability rather than its instruction-following capability.

\section{Failure Case}
\begin{figure*}[t]
\centering

\errorcase{}{How many elephants are there in the image?}{5}{4}
\hfill
\errorcase{}{How many plates are there in the image?}{12}{11}
\hfill
\errorcase{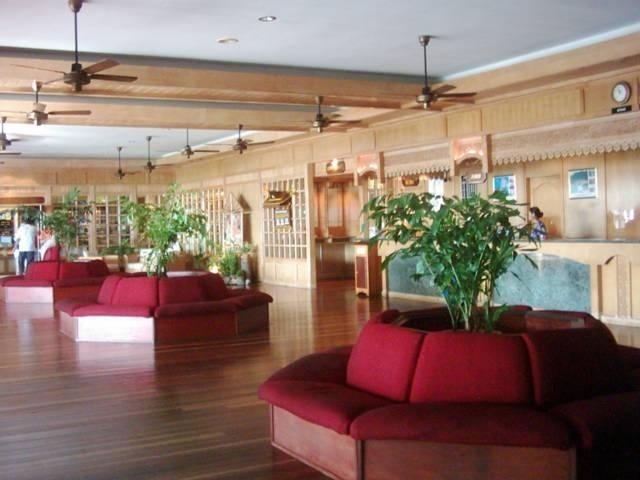}{How many red sofas are visible in the lobby?}{3}{6}
\hfill
\errorcase{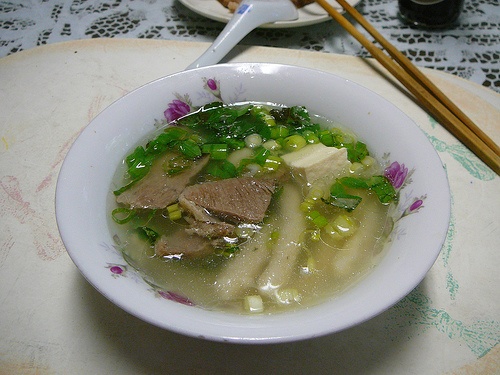}{How many purple flowers are printed on the rim of the bowl?}{6}{4}

\vspace{1pt}
\makebox[0.48\linewidth]{\scriptsize\textbf{Atomic Counting}}\hfill\makebox[0.48\linewidth]{\scriptsize\textbf{Chromatic Attribution}}

\vspace{6pt}
\errorcase{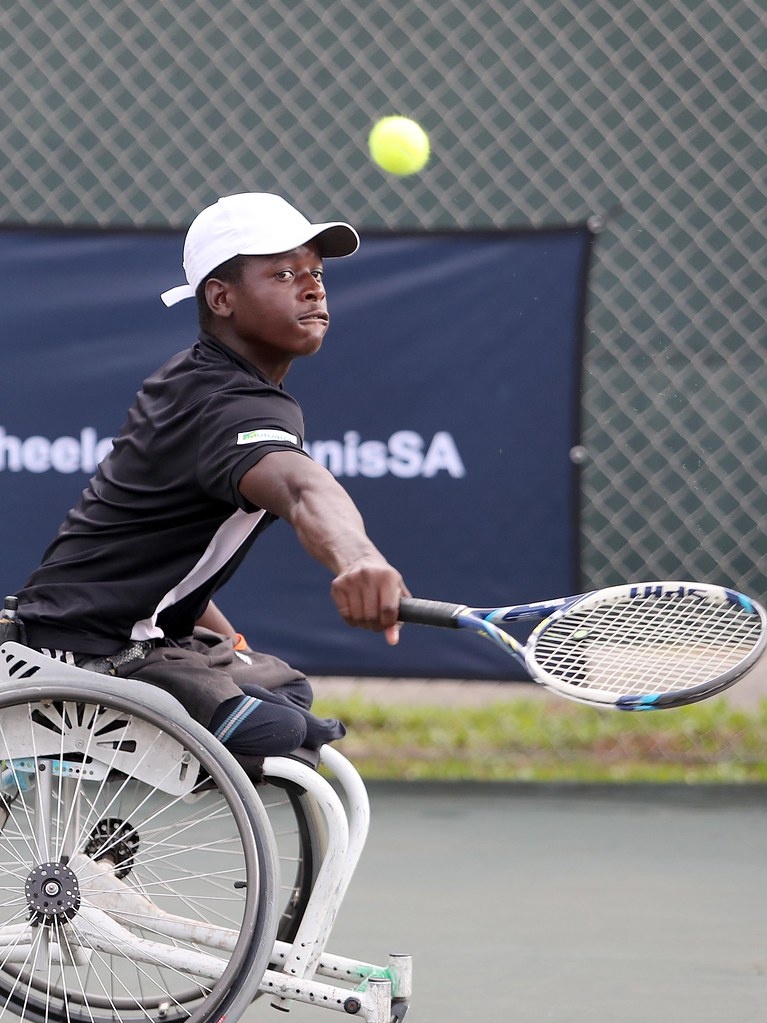}{How many mesh fences are in the background?}{1}{5}
\hfill
\errorcase{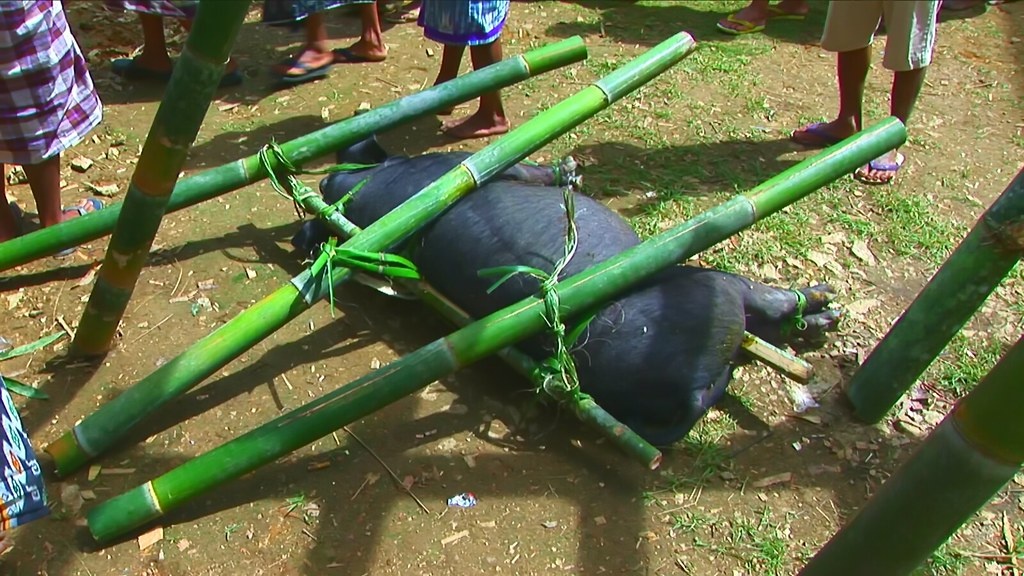}{How many bamboo poles are visible in total?}{7}{6}
\hfill
\errorcase{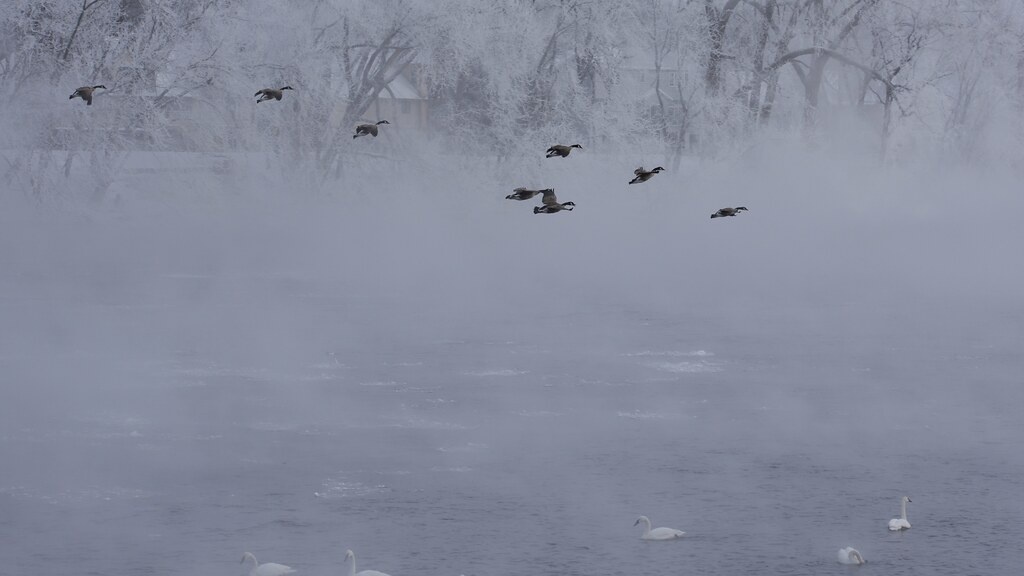}{How many large birds are visible?}{8}{13}
\hfill
\errorcase{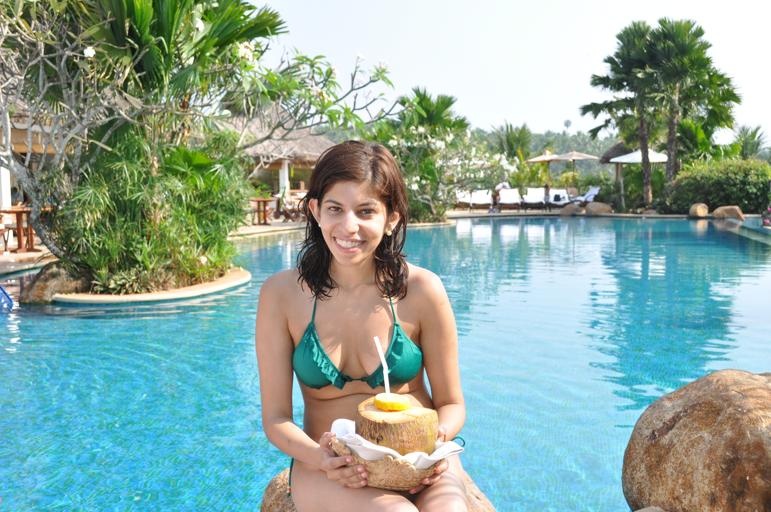}{How many large rocks are in the foreground?}{2}{3}

\vspace{1pt}
\makebox[0.48\linewidth]{\scriptsize\textbf{Material Attribution}}\hfill\makebox[0.48\linewidth]{\scriptsize\textbf{Scale Attribution}}

\vspace{6pt}
\errorcase{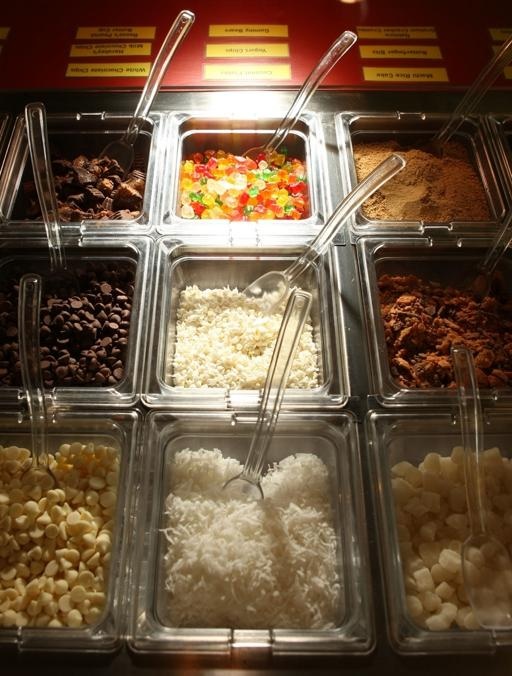}{How many spoons have handles pointing towards the top right?}{6}{3}
\hfill
\errorcase{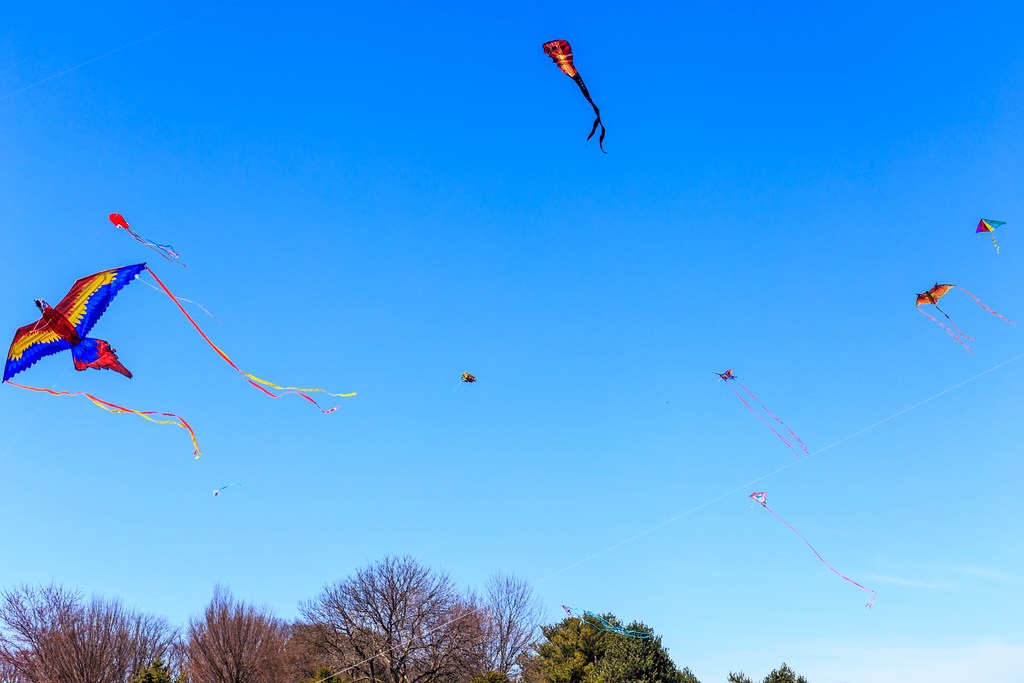}{How many kites are flying in the sky?}{10}{8}
\hfill
\errorcase{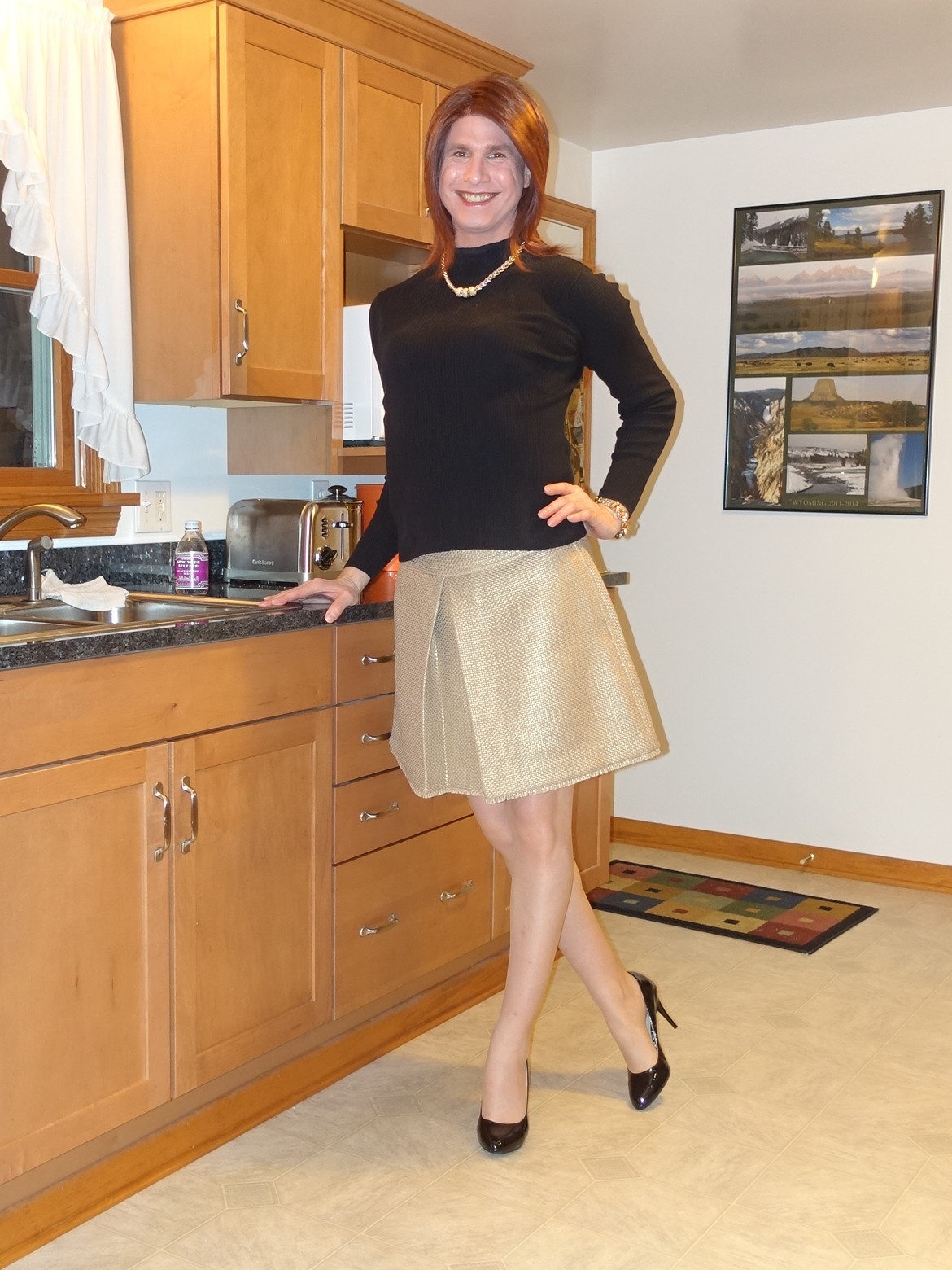}{How many square patches are visible on the rug?}{15}{11}
\hfill
\errorcase{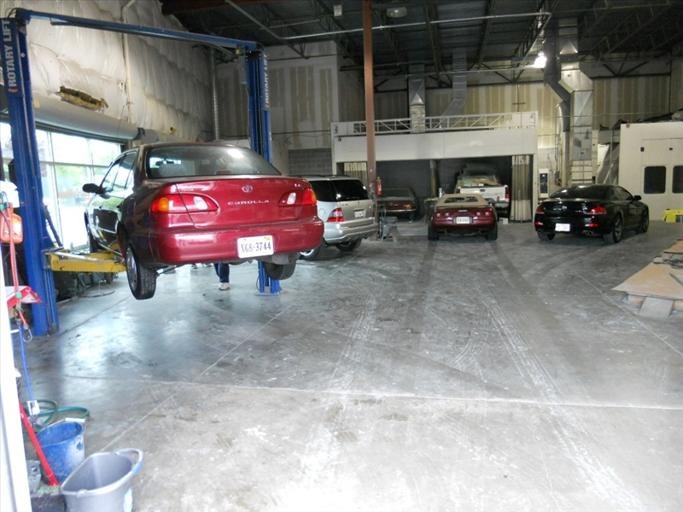}{How many cars are facing away from the camera?}{6}{5}

\vspace{1pt}
\makebox[0.48\linewidth]{\scriptsize\textbf{Action \& State Attribution}}\hfill\makebox[0.48\linewidth]{\scriptsize\textbf{General Semantic Attribution}}

\caption{Qualitative error examples of Gemini-3.1-Pro-Preview on \textbf{Semantic Counting}. \textcolor{correct}{Green}: ground truth. \textcolor{wrong}{Red}: model prediction.}
\label{fig:qual_semantic}
\end{figure*}

\begin{figure*}[t]
\centering

\errorcase{}{How many shirts are there in the red rectangle?}{2}{1}
\hfill
\errorcase{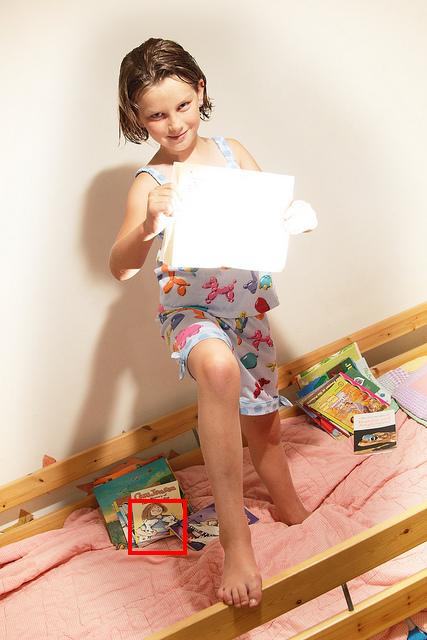}{How many dogs are there in the red rectangle?}{2}{1}
\hfill
\errorcase{}{How many wine bottles are in <bbox>248 398 468 625</bbox>?}{4}{0}
\hfill
\errorcase{}{How many monitors are in <bbox>45 208 358 602</bbox>?}{2}{0}

\vspace{1pt}
\makebox[0.48\linewidth]{\scriptsize\textbf{Visual-Prompt Region Grounding}}\hfill\makebox[0.48\linewidth]{\scriptsize\textbf{Coordinate-Prompt Region Grounding}}

\vspace{6pt}
\errorcase{}{How many wheels are there in right half of the image?}{2}{4}
\hfill
\errorcase{}{How many forks are there in left half of the image?}{2}{1}
\hfill
\errorcase{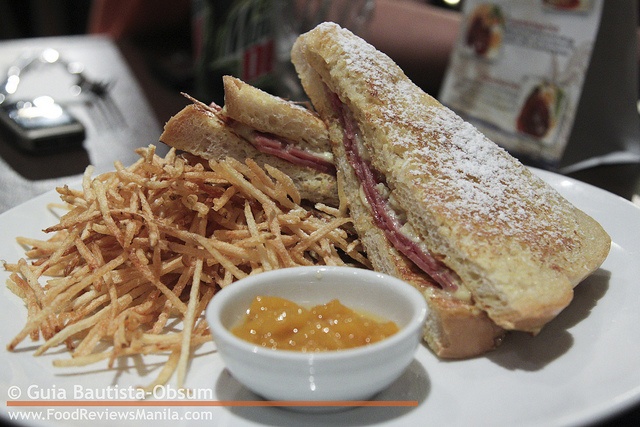}{How many more sandwich triangles than small sauce bowls?}{1}{2}
\hfill
\errorcase{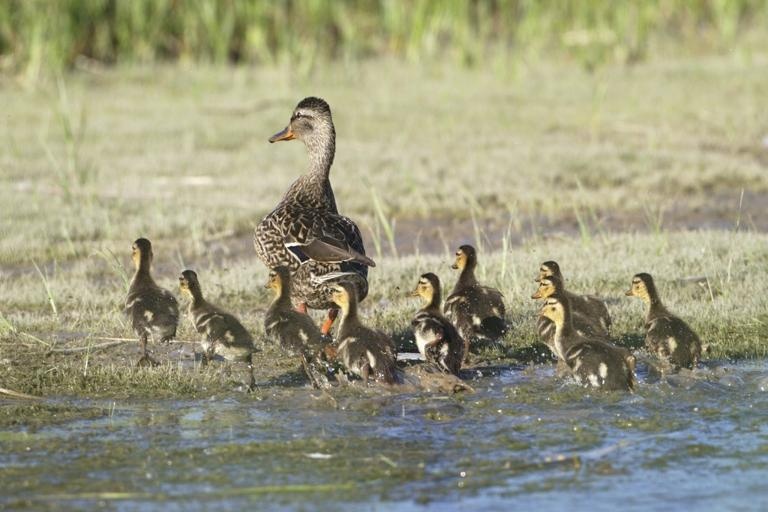}{How many more ducklings than adult ducks?}{9}{8}

\vspace{1pt}
\makebox[0.48\linewidth]{\scriptsize\textbf{Relative Canonical Orientation}}\hfill\makebox[0.48\linewidth]{\scriptsize\textbf{Differential Comparison}}

\vspace{6pt}
\errorcase{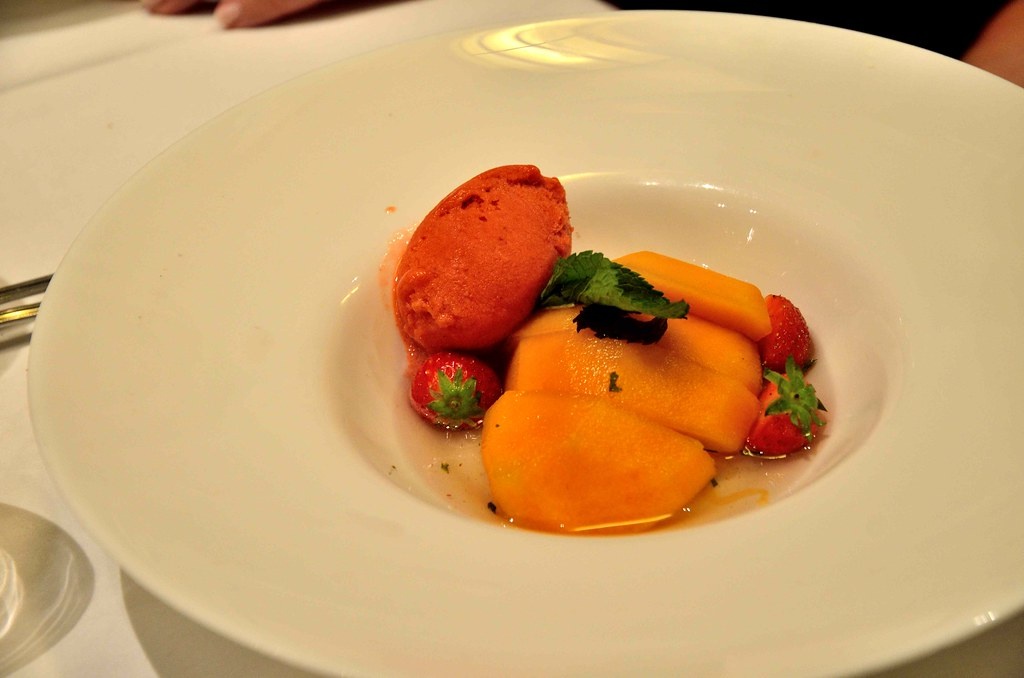}{Total number of strawberries and orange fruit slices combined?}{7}{6}
\hfill
\errorcase{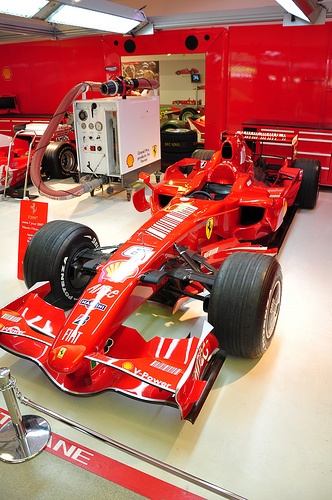}{How many large rectangular ceiling light fixtures in total?}{3}{2}
\hfill
\errorcase{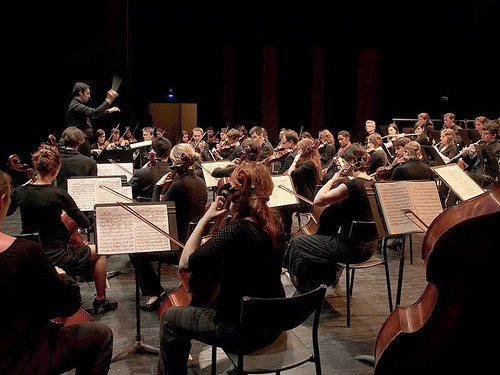}{Excluding the conductor, how many people are standing?}{0}{21}
\hfill
\errorcase{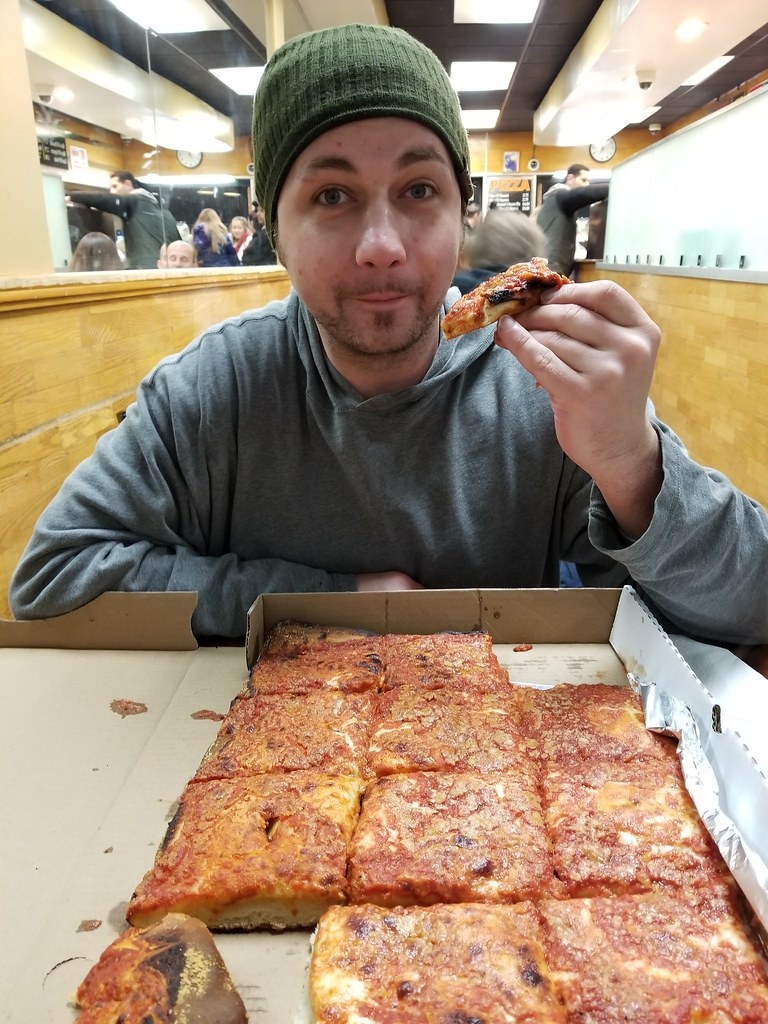}{Excluding the slice in the man's hand, how many pizza pieces are in the box?}{11}{10}

\vspace{1pt}
\makebox[0.48\linewidth]{\scriptsize\textbf{Joint-Set Aggregation}}\hfill\makebox[0.48\linewidth]{\scriptsize\textbf{Complementary Exclusion}}

\vspace{6pt}
\errorcase{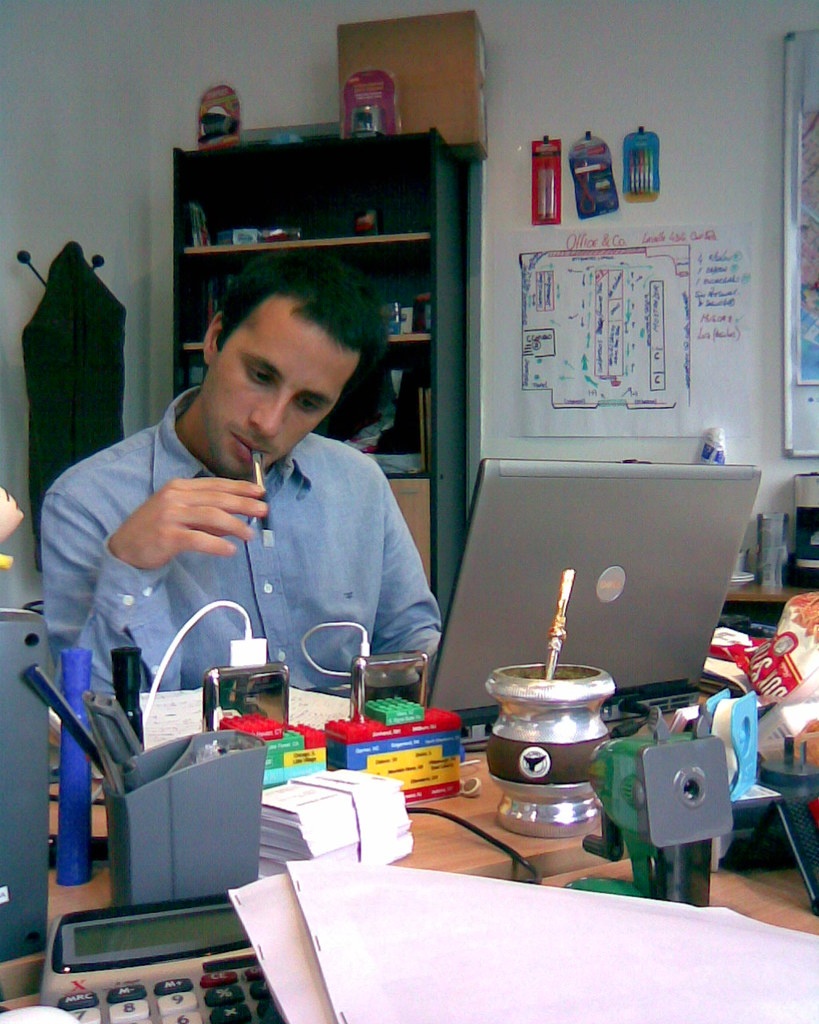}{How many distinct colors in the building blocks?}{5}{4}
\hfill
\errorcase{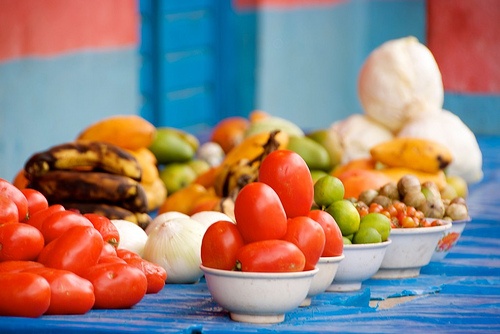}{How many different types of produce are in the small white bowls?}{4}{3}
\hfill
\begin{minipage}[t]{0.235\linewidth}\end{minipage}%
\hfill
\begin{minipage}[t]{0.235\linewidth}\end{minipage}%

\vspace{1pt}
\makebox[0.48\linewidth]{\scriptsize\textbf{Categorical Cardinality}}\hfill\makebox[0.48\linewidth]{}

\caption{Qualitative error examples of Gemini-3.1-Pro-Preview on \textbf{Analytical Counting}. \textcolor{correct}{Green}: ground truth. \textcolor{wrong}{Red}: model prediction. }
\label{fig:qual_analytic}
\end{figure*}

\begin{figure*}[t]
\centering

\errorcase{}{How many turtlenecks are there in the image?}{0}{6}
\hfill
\errorcase{}{How many parachutes are there in the image?}{0}{5}
\hfill
\errorcase{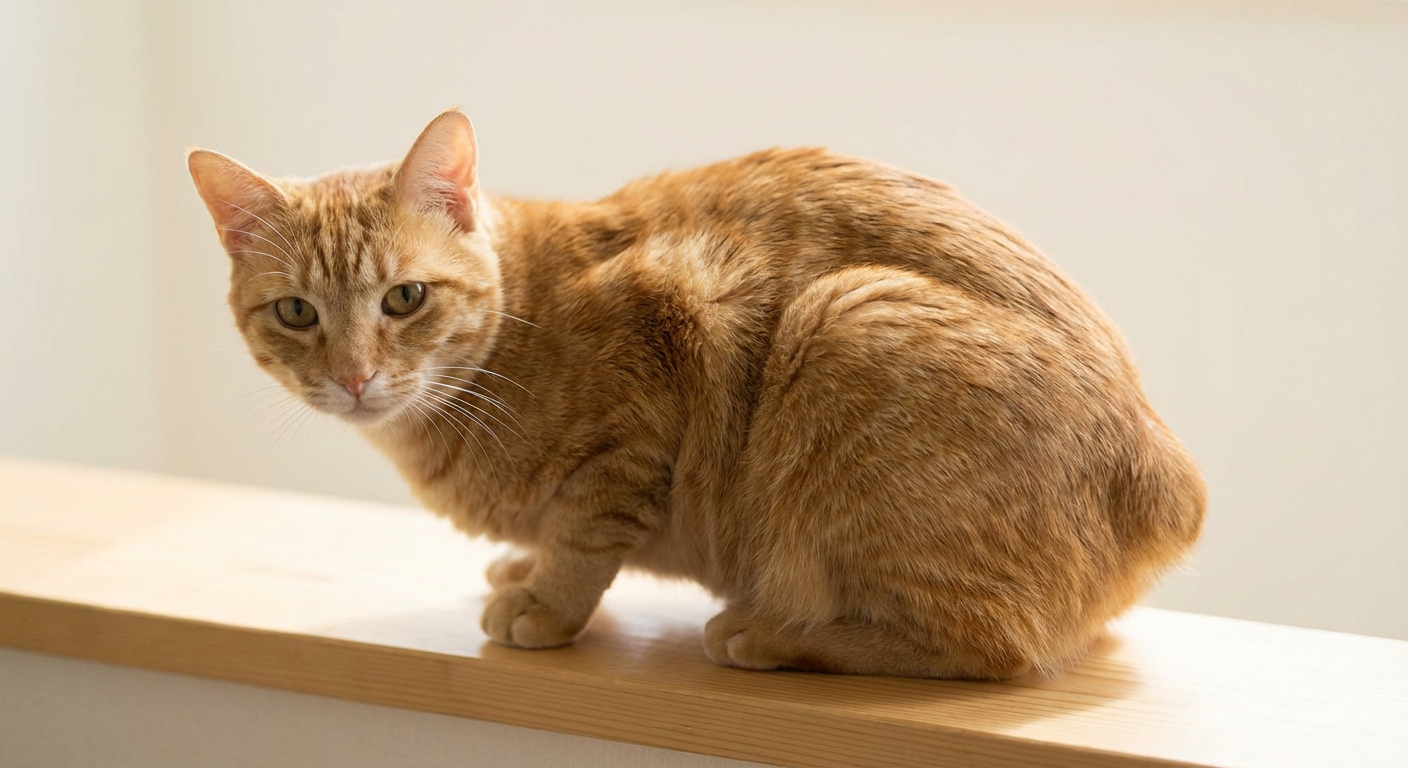}{How many tails does the cat in the image have?}{0}{1}
\hfill
\errorcase{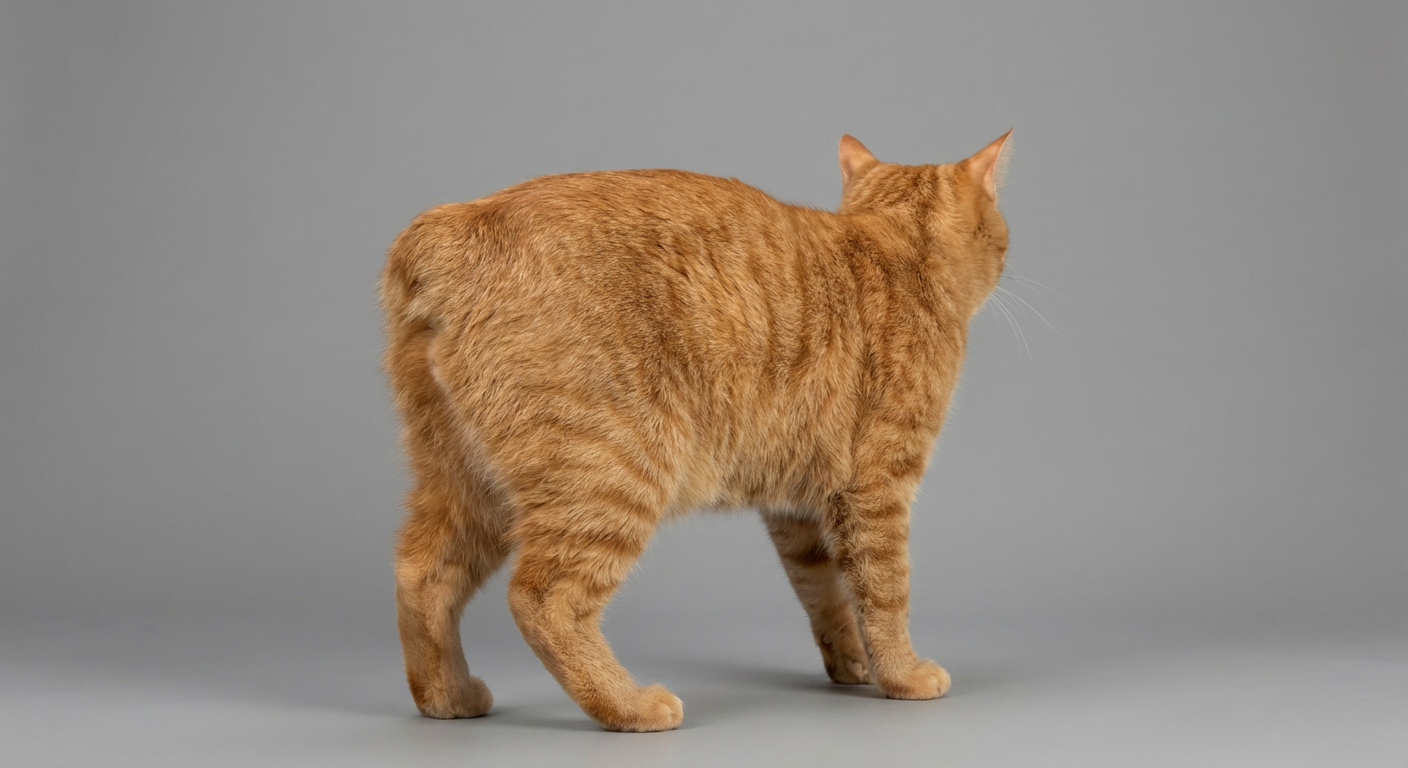}{How many tails does the cat in the image have?}{0}{1}

\vspace{1pt}
\makebox[0.48\linewidth]{\scriptsize\textbf{Null-Target Prompting}}\hfill\makebox[0.48\linewidth]{\scriptsize\textbf{Linguistic Prior Conflict}}

\vspace{6pt}
\errorcase{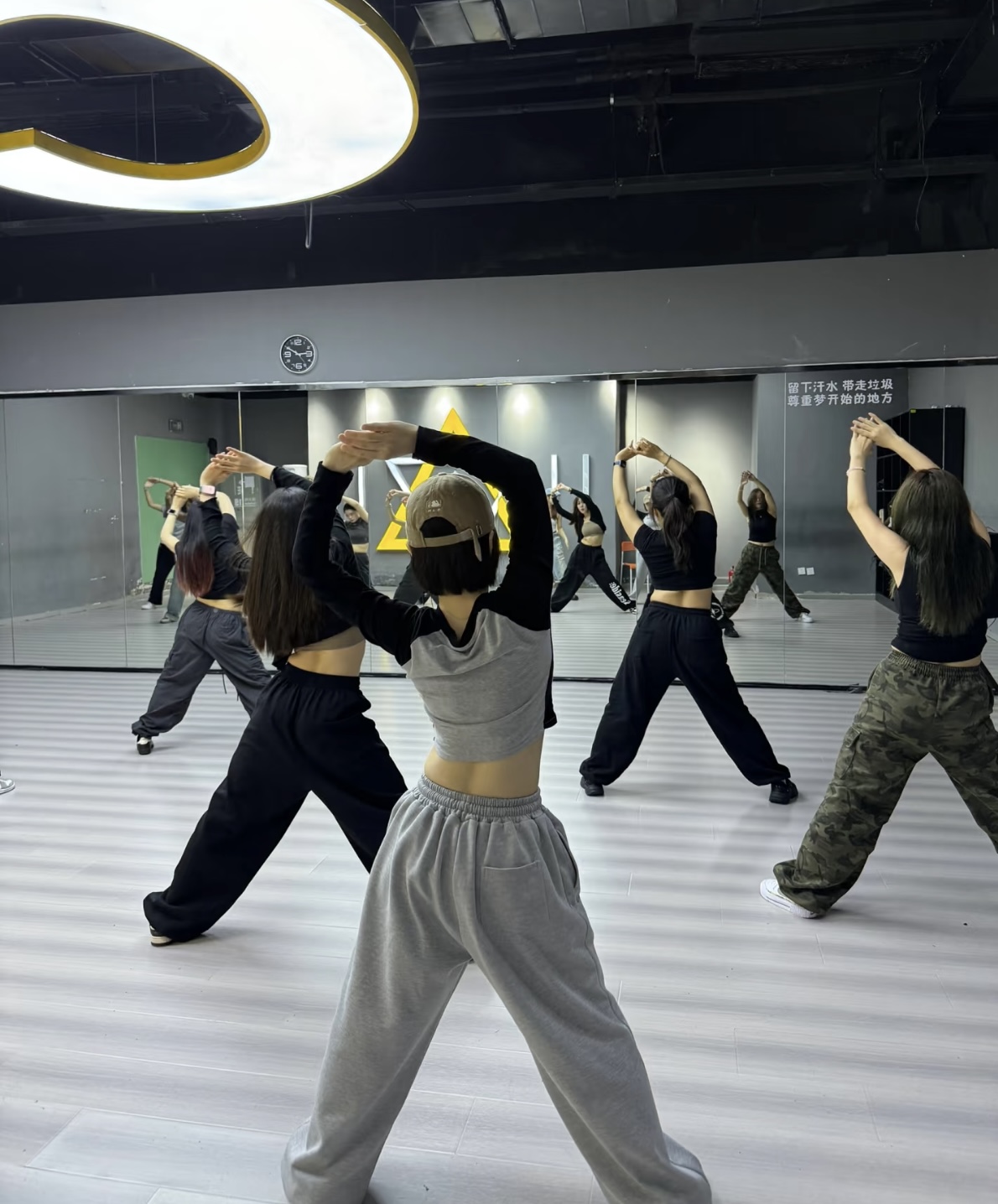}{How many people are there in the picture?}{5}{11}
\hfill
\errorcase{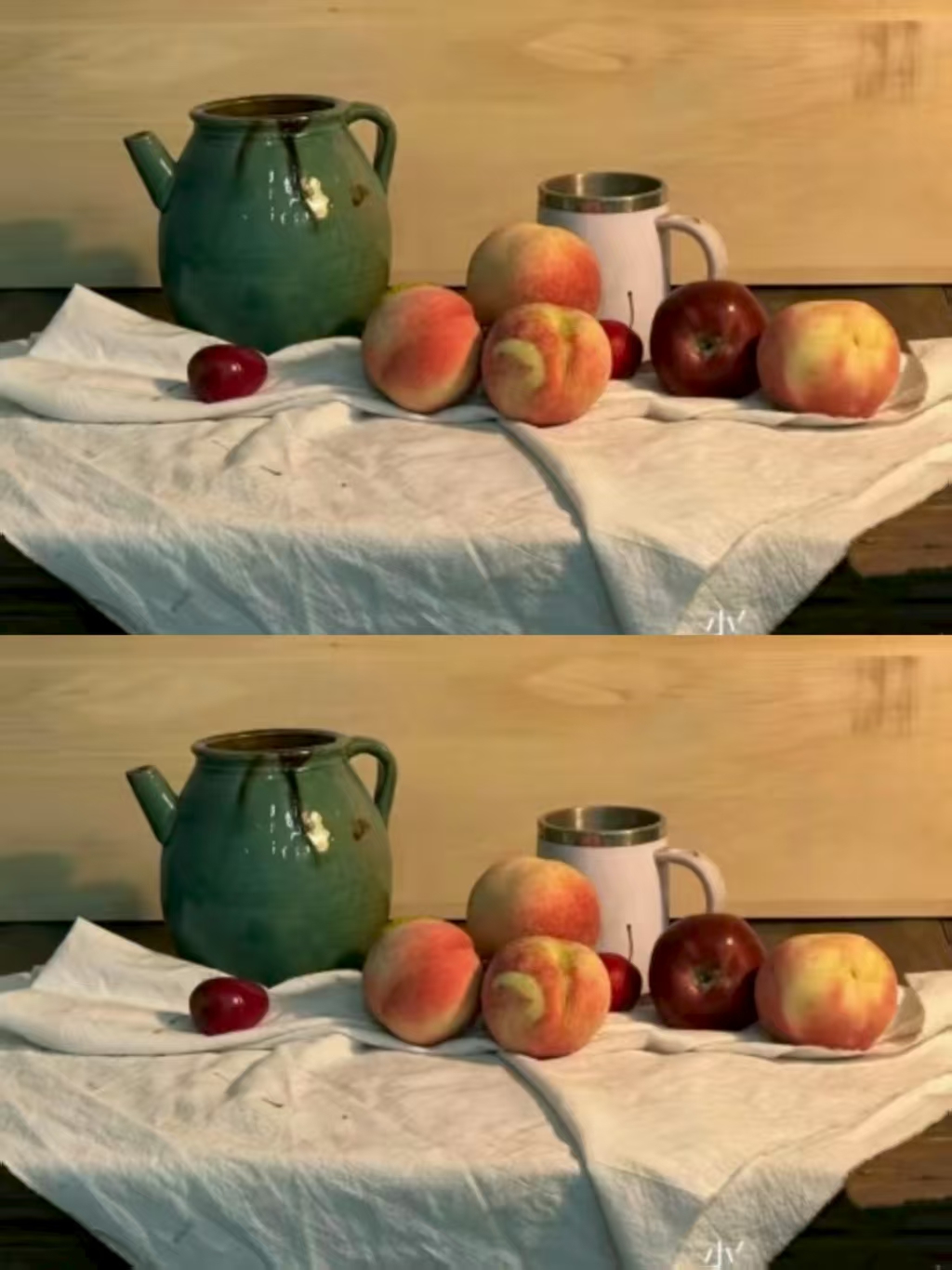}{How many real peaches are there in the picture?}{8}{4}
\hfill
\errorcase{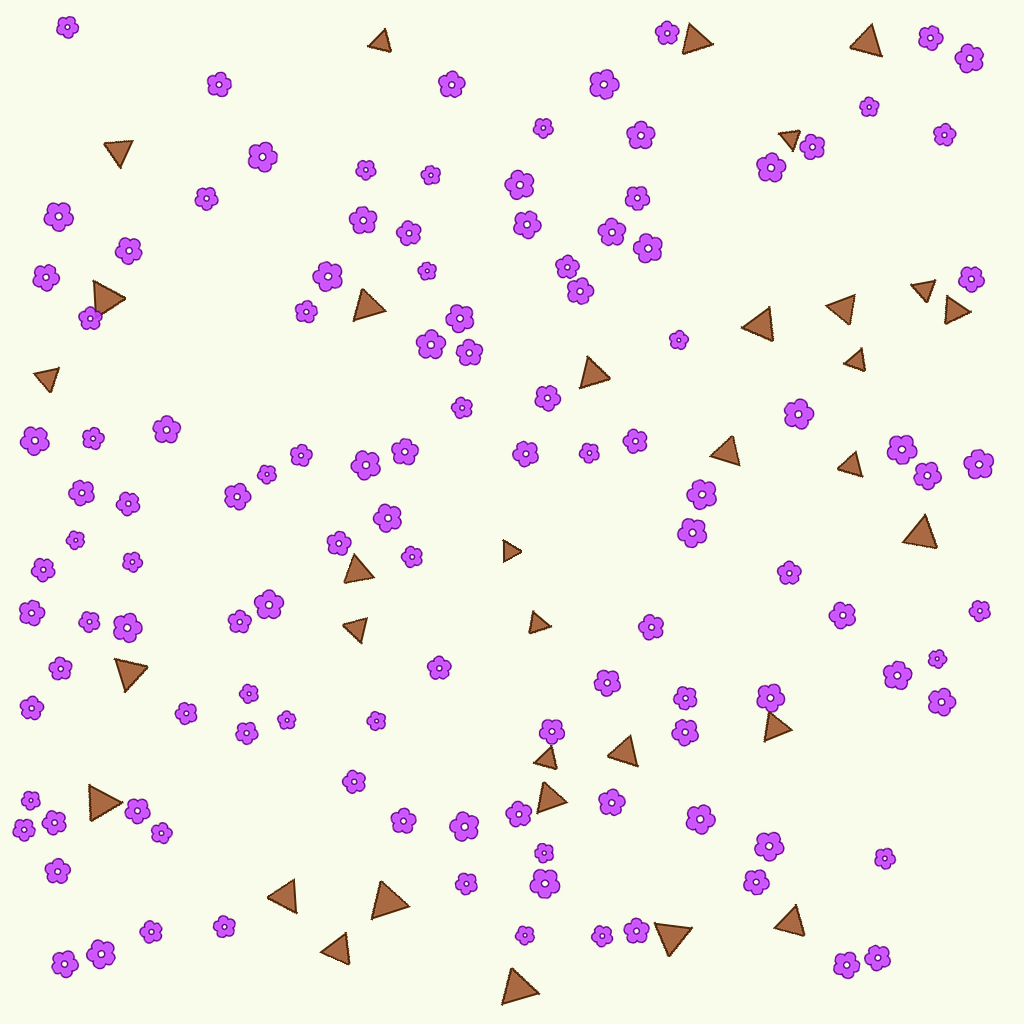}{How many flowers are in this image?}{117}{115}
\hfill
\errorcase{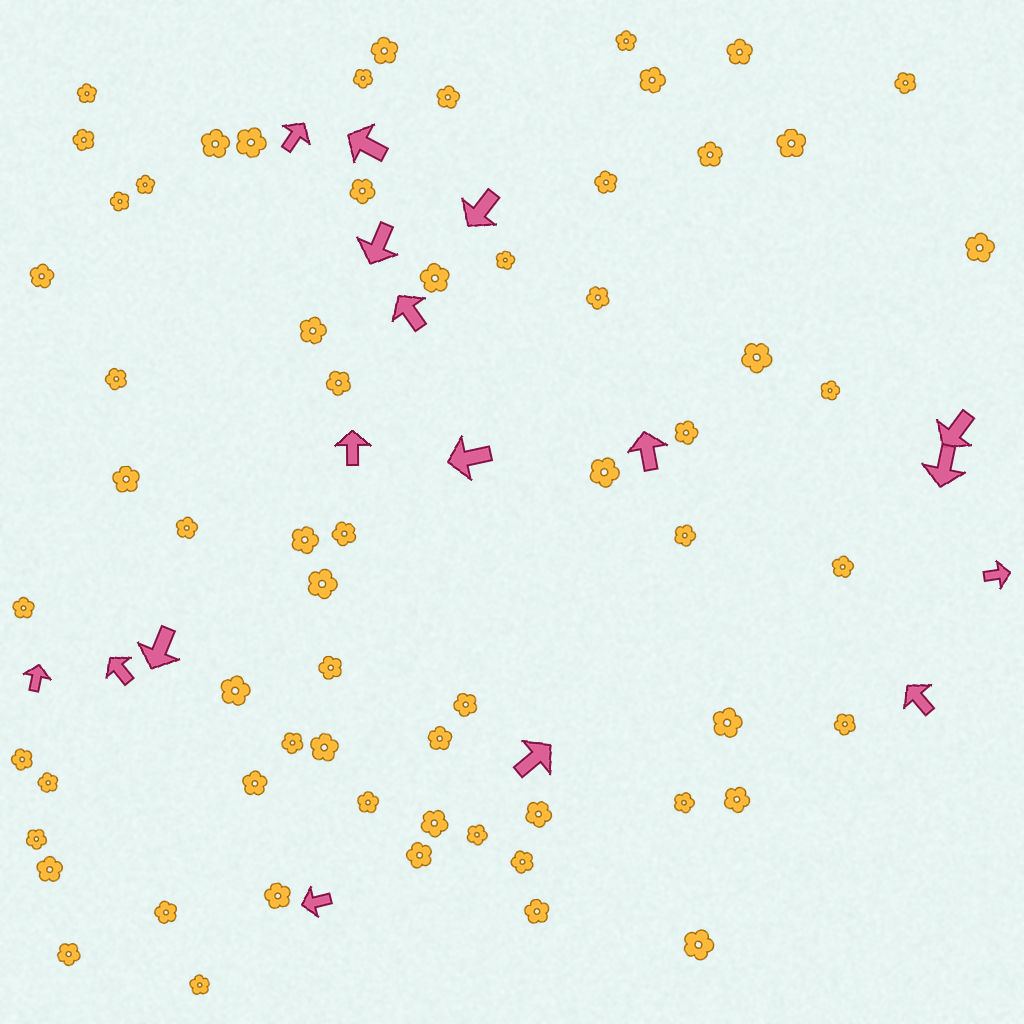}{How many flowers are in this image?}{64}{63}

\vspace{1pt}
\makebox[0.48\linewidth]{\scriptsize\textbf{Visual Distractor Illusion}}\hfill\makebox[0.48\linewidth]{\scriptsize\textbf{High-Density Enumeration}}

\vspace{6pt}
\errorcase{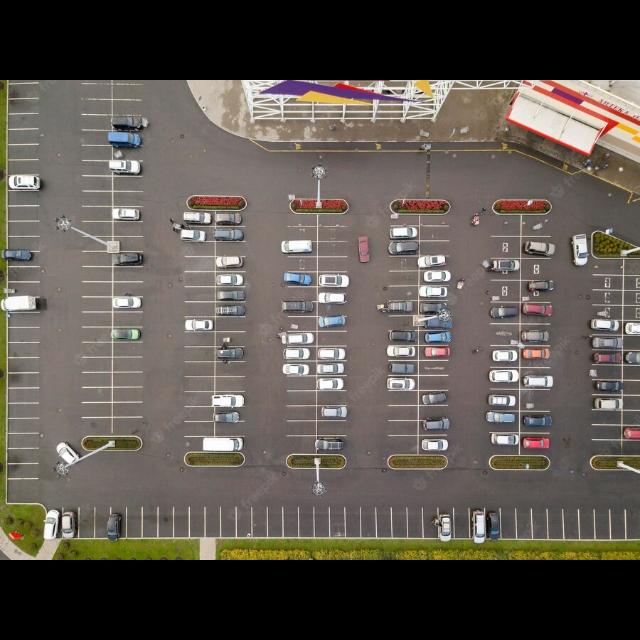}{How many cars are there in the image?}{84}{86}
\hfill
\errorcase{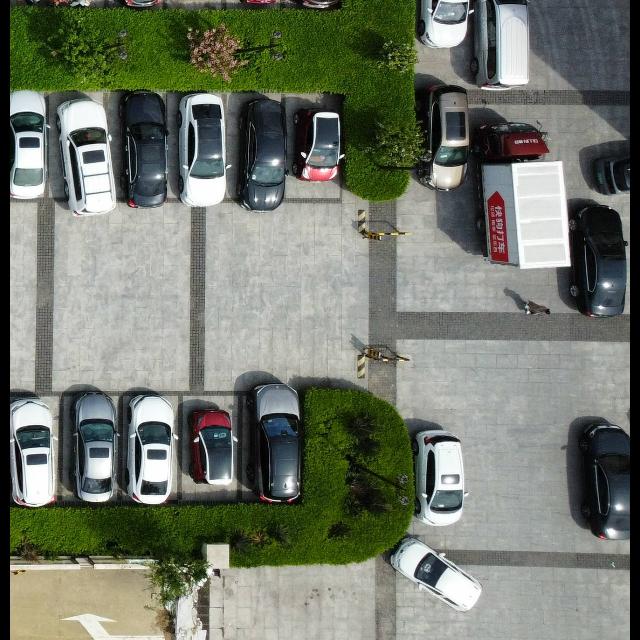}{How many cars are there in the image?}{19}{21}
\hfill
\errorcase{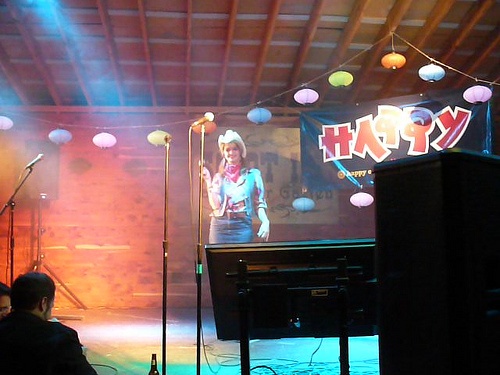}{How many lanterns are there in the image?}{13}{10}
\hfill
\errorcase{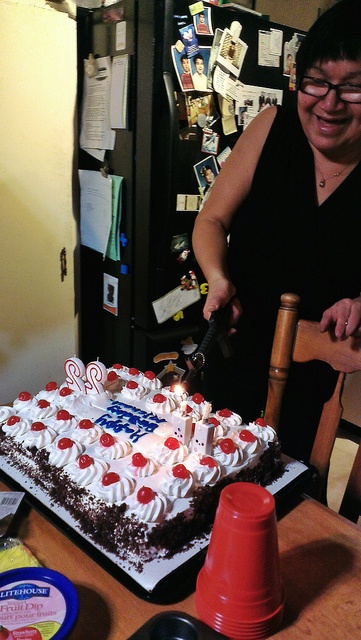}{How many cherries are there in the image?}{31}{28}

\vspace{1pt}
\makebox[0.48\linewidth]{\scriptsize\textbf{Aerial Perspective Shift}}\hfill\makebox[0.48\linewidth]{\scriptsize\textbf{Small-Scale Enumeration}}

\vspace{6pt}
\errorcase{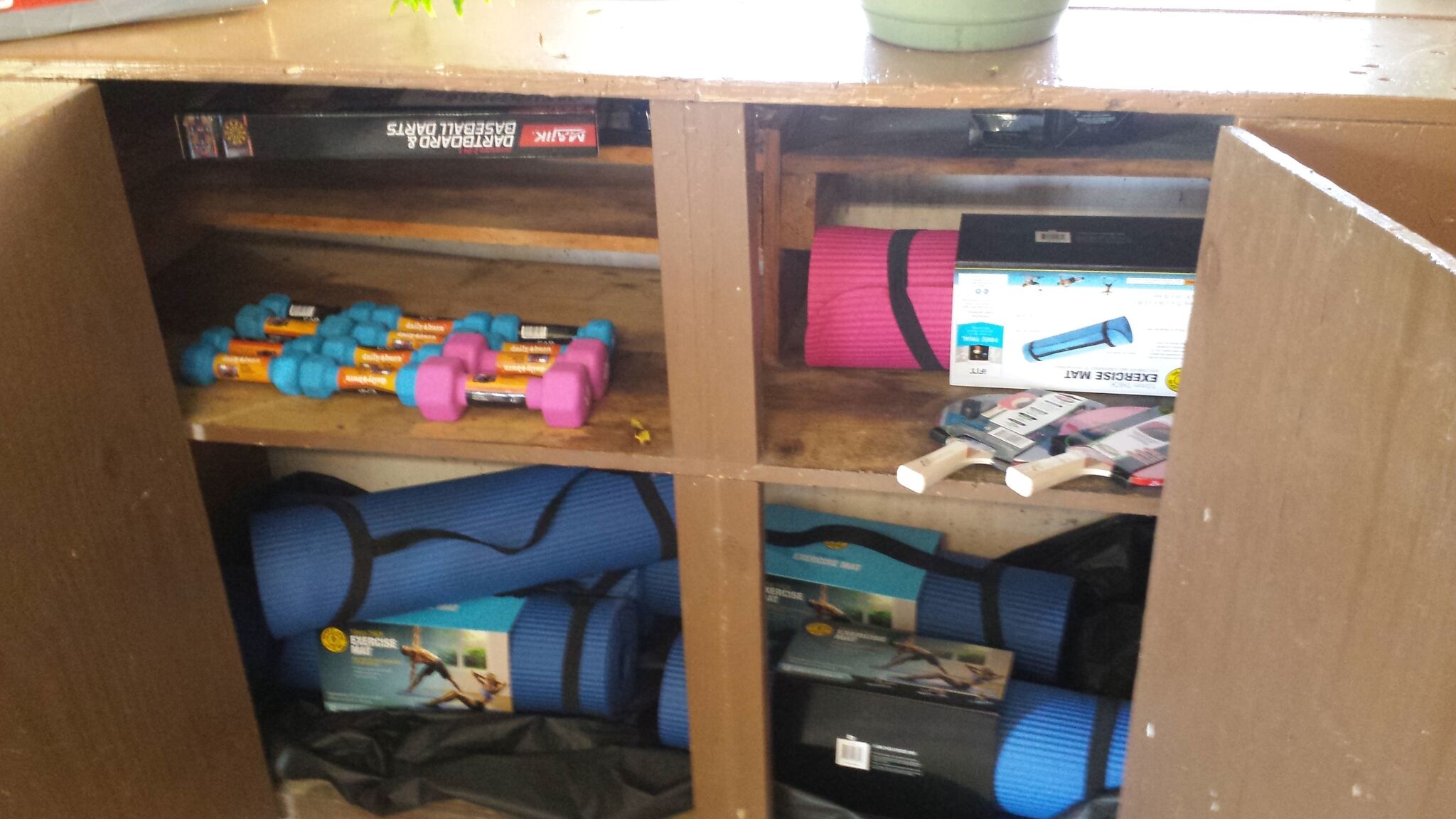}{How many dumbbells are there in the image?}{12}{9}
\hfill
\errorcase{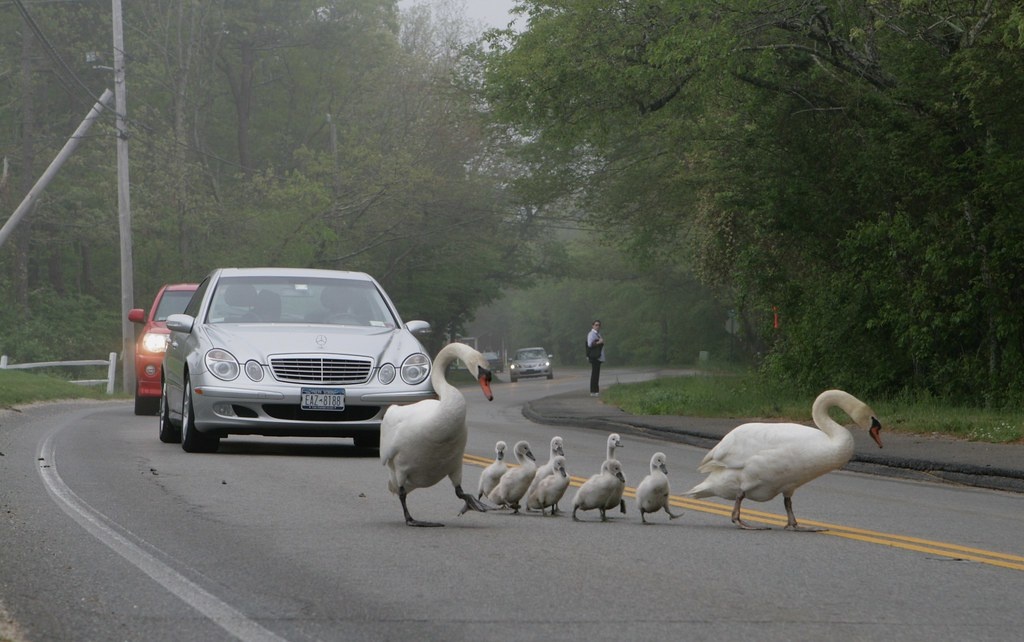}{How many geese are there in the image?}{9}{7}
\hfill
\begin{minipage}[t]{0.235\linewidth}\end{minipage}%
\hfill
\begin{minipage}[t]{0.235\linewidth}\end{minipage}%

\vspace{1pt}
\makebox[0.48\linewidth]{\scriptsize\textbf{Partial Object Occlusion}}\hfill\makebox[0.48\linewidth]{}

\caption{Qualitative error examples of Gemini-3.1-Pro-Preview on \textbf{Robustness Testing}. \textcolor{correct}{Green}: ground truth. \textcolor{wrong}{Red}: model prediction. The model hallucinates objects that are absent, adheres to linguistic priors over visual evidence, and systematically undercounts under occlusion or small scale.}
\label{fig:qual_robustness}
\end{figure*}

We provide qualitative failure cases of Gemini-3.1-Pro across all subsets of HoloCount in Figures~\ref{fig:qual_semantic},~\ref{fig:qual_analytic}, and~\ref{fig:qual_robustness}. These examples illustrate representative error patterns including attribute misinterpretation, spatial grounding failure, logical reasoning breakdown, language prior conflict, and perception degradation under adverse conditions.

\section{Future Work}

In this work, we present HoloCount, a comprehensive benchmark for evaluating the counting capabilities of multimodal large language models across semantic counting, analytic counting, and robustness testing. Our evaluation reveals significant gaps in current models, particularly in analytical reasoning, small-scale enumeration, and resistance to linguistic priors. Building on these findings, we identify several promising directions for future research:

\noindent \textbf{Video Counting.} HoloCount currently focuses on static images. However, real-world counting often occurs in dynamic environments where objects enter, exit, or become temporarily occluded across frames. Extending the benchmark to video sequences would enable evaluation of temporal counting abilities, such as tracking cumulative arrivals, counting unique individuals across frames, and maintaining accurate counts despite motion blur or transient occlusions.

\noindent\textbf{Grounded Counting with Localization.} Our current evaluation only requires models to output a single integer. A natural extension is to require models to jointly count and localize --- producing not only the total count but also the spatial location (e.g., bounding boxes or points) of each counted instance. This would enable deeper diagnosis of counting errors: whether a model counts incorrectly because it misidentifies objects, double-counts the same instance, or fails to detect certain targets entirely.

\noindent\textbf{Multi-step Compositional Counting.} While HoloCount includes basic analytic counting (comparison, aggregation, exclusion), future work can explore more complex compositional chains that require multiple reasoning steps. For example, questions like ``How many objects satisfy condition A but not condition B, and are larger than the average size of category C?'' demand sequential constraint evaluation, arithmetic operations, and cross-category comparison --- pushing models toward genuine multi-step visual reasoning rather than pattern-matched responses.

\noindent\textbf{Counting with World Knowledge.} Current counting tasks define targets by visual attributes or explicit category names. A cognitively richer direction involves counting that requires external world knowledge. For instance, ``How many mammals are in this image?'' requires taxonomic knowledge to classify each visible animal, while ``How many edible items are on the table?'' demands functional knowledge about object affordances. This direction bridges visual counting with knowledge-grounded reasoning, testing whether models can integrate encyclopedic knowledge with perceptual evidence.




\end{document}